\documentclass[10pt,twocolumn,letterpaper]{article}

\usepackage[pagenumbers]{cvpr} 

\usepackage{graphicx}
\usepackage{amsmath}
\usepackage{amssymb}
\usepackage{booktabs}
\usepackage{tabularx}
\usepackage{multirow}
\usepackage{flushend}
\usepackage{mathtools}

\usepackage[pagebackref,breaklinks,colorlinks]{hyperref}

\usepackage[capitalize]{cleveref}
\crefname{section}{Sec.}{Secs.}
\Crefname{section}{Section}{Sections}
\Crefname{table}{Table}{Tables}
\crefname{table}{Tab.}{Tabs.}

\newcommand{\pardev}[2]{\frac{\partial {#1}}{\partial {#2}}}

\newcommand{\netparam}{\sigma}

\newcommand{\function}[1]{{#1}_{\netparam_{#1}}}

\newcommand{\loss}{\mathcal{L}}

\newcommand{\boldparagraph}[1]{\vspace{0.2cm}\noindent{\bf #1:}}

\def\cvprPaperID{6545} 
\def\confName{CVPR}
\def\confYear{2022}

\usepackage{overpic}
\usepackage{enumitem} %
\usepackage{overpic} %
\usepackage{color}
\usepackage[dvipsnames]{xcolor}
\usepackage{paralist}
\usepackage{soul} %

\definecolor{turquoise}{cmyk}{0.65,0,0.1,0.3}
\definecolor{purple}{rgb}{0.65,0,0.65}
\definecolor{dark_green}{rgb}{0, 0.5, 0}
\definecolor{orange}{rgb}{0.8, 0.6, 0.2}
\definecolor{red}{rgb}{0.8, 0.2, 0.2}
\definecolor{darkred}{rgb}{0.6, 0.1, 0.05}
\definecolor{blueish}{rgb}{0.0, 0.3, .6}
\definecolor{light_gray}{rgb}{0.7, 0.7, .7}
\definecolor{pink}{rgb}{1, 0, 1}
\definecolor{greyblue}{rgb}{0.25, 0.25, 1}

\newif\ifshowcomments
\showcommentsfalse %

\ifshowcomments
    \newcommand{\todo}[1]{{\color{red}#1}}
    \newcommand{\TODO}[1]{\textbf{\color{red}[TODO: #1]}}
    \newcommand{\yf}[1]{{\color{blueish}#1}} %
    \newcommand{\YF}[1]{{\color{blueish}{\bf [YF: #1]}}} %
    \newcommand{\Yf}[1]{\marginpar{\tiny{\textcolor{blueish}{#1}}}} %
    \newcommand{\va}[1]{{\color{purple}#1}} %
    \newcommand{\VA}[1]{{\color{purple}{\bf [VA: #1]}}} %
    \newcommand{\Va}[1]{\marginpar{\tiny{\textcolor{purple}{#1}}}} %
    \newcommand{\mb}[1]{{\color{dark_green}#1}} %
    \newcommand{\MB}[1]{{\color{dark_green}{\bf [MCB: #1]}}} %
    \newcommand{\Mb}[1]{\marginpar{\tiny{\textcolor{dark_green}{#1}}}} %
    \newcommand{\xc}[1]{{\color{darkred}#1}} %
    \newcommand{\XC}[1]{{\color{darkred}{\bf [XC: #1]}}} %
    \newcommand{\Xc}[1]{\marginpar{\tiny{\textcolor{darkred}{#1}}}} %
    \newcommand{\oh}[1]{{\color{orange}#1}} %
    \newcommand{\OH}[1]{{\color{orange}{\bf [OH: #1]}}} %
    \newcommand{\Oh}[1]{\marginpar{\tiny{\textcolor{orange}{#1}}}} %
    \newcommand{\mcb}[1]{{\color{greyblue}#1}} %
    \newcommand{\MCB}[1]{{\color{greyblue}{\bf [MCB: #1]}}} %
    \newcommand{\Mcb}[1]{\marginpar{\tiny{\textcolor{greyblue}{#1}}}} %
\else
    \newcommand{\todo}[1]{\unskip}
    \newcommand{\TODO}[1]{\unskip}
    \newcommand{\yf}[1]{\unskip} %
    \newcommand{\YF}[1]{\unskip} %
    \newcommand{\Yf}[1]{\unskip} %
    \newcommand{\va}[1]{\unskip} %
    \newcommand{\VA}[1]{\unskip} %
    \newcommand{\Va}[1]{\unskip} %
    \newcommand{\mb}[1]{\unskip} %
    \newcommand{\MB}[1]{\unskip} %
    \newcommand{\Mb}[1]{\unskip} %
    \newcommand{\xc}[1]{\unskip} %
    \newcommand{\XC}[1]{\unskip} %
    \newcommand{\Xc}[1]{\unskip} %
    \newcommand{\oh}[1]{\unskip} %
    \newcommand{\OH}[1]{\unskip} %
    \newcommand{\Oh}[1]{\unskip} %
    \newcommand{\mcb}[1]{\unskip} %
    \newcommand{\MCB}[1]{\unskip} %
    \newcommand{\Mcb}[1]{\unskip}%
\fi

\usepackage{blindtext}

\renewcommand{\paragraph}[1]{\vspace{1em}\noindent\textbf{#1}.}

\newcommand{\methodname}{IMavatar\xspace}
\newcommand{\suppmat}{Sup.~Mat\xspace}
\begin{document}
\title{I M Avatar: Implicit Morphable Head Avatars from Videos}

\author{Yufeng Zheng$^{1,3}$\quad Victoria Fernández Abrevaya$^{2}$\quad Marcel C. Bühler$^{1}$\quad Xu Chen$^{1,3}$\\ Michael J. Black$^{2}$\quad Otmar Hilliges$^{1}$\\
$^1$ETH Zürich\quad $^2$Max Planck Institute for Intelligent Systems, Tübingen \\ $^3$Max Planck ETH Center for Learning Systems \\
{}
}

\maketitle
\begin{abstract}
Traditional 3D morphable face models (3DMMs) provide fine-grained control over expression but cannot easily capture geometric and appearance details.
Neural volumetric representations approach photorealism but are hard to animate and do not generalize well to unseen expressions. 
To tackle this problem, we propose \methodname{} (\textbf{I}mplicit \textbf{M}orphable \textbf{avatar}), a novel method for learning implicit head avatars from monocular videos.
Inspired by the fine-grained control mechanisms afforded by conventional 3DMMs, we represent the expression- and pose-related deformations via learned blendshapes and skinning fields.
These attributes are pose-independent and can be used to morph the canonical geometry and texture fields given novel expression and pose parameters. 
We employ ray marching and iterative root-finding to locate the canonical surface intersection for each pixel.
A key contribution is our novel analytical gradient formulation that enables end-to-end training of \methodname{}s from videos. 
We show quantitatively and qualitatively that our method improves geometry and covers a more complete expression space compared to state-of-the-art methods. Code and data can be found at %
\href{https://ait.ethz.ch/projects/2022/IMavatar/}{https://ait.ethz.ch/projects/2022/IMavatar/}.

\end{abstract}
\setlength{\topsep}{0pt}
\setlength{\parskip}{.5ex}
\renewcommand{\floatsep}{1ex}
\renewcommand{\textfloatsep}{1ex}
\renewcommand{\dblfloatsep}{1ex}
\renewcommand{\dbltextfloatsep}{1ex}
\section{Introduction}
\label{sec:intro}

\begin{figure}[t]
\centerline{
\includegraphics[trim=0 4em 0 0, clip=true, width=\linewidth]{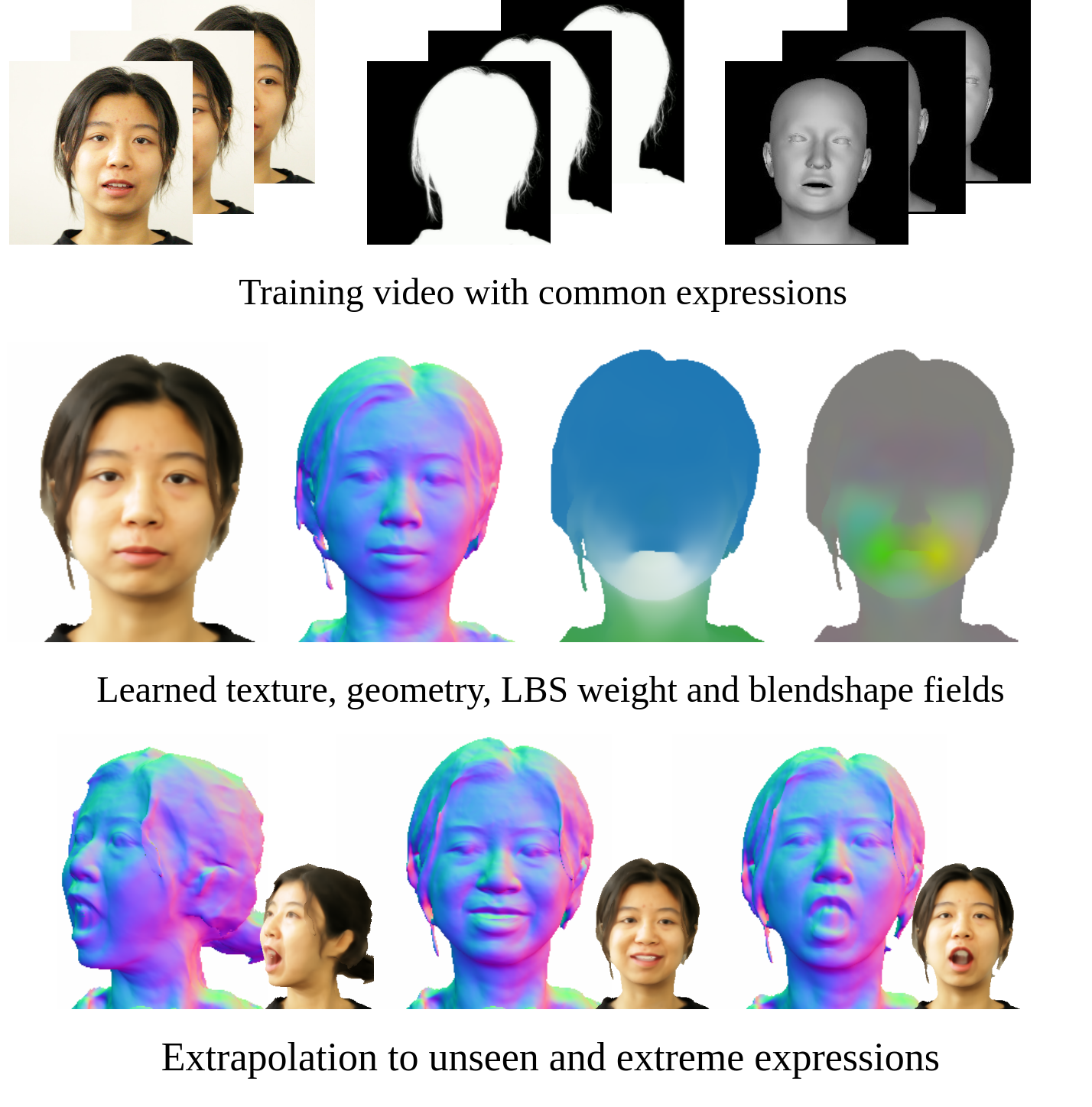}}
\caption{
\textbf{IMavatar.} We learn personal avatars from RGB videos, represented by  canonical shape, texture, and deformation fields. An implicit morphing formulation allows generalization to unseen expressions and poses outside the training distribution.
}
\label{fig:teaser}
\end{figure}
Methods to automatically create animatable personal avatars in unobtrusive and readily available settings (i.e., from monocular videos) have many applications in VR/AR games and telepresence. 
Such applications require faithful renderings of the deforming facial geometry and expressions, detailed facial appearance and accurate reconstruction of the entire head and hair region. 
Conventional methods \cite{Deng19, Garrido16, Gecer19, Romdhani05, Sanyal19, Shang20, Tewari19, Feng21} based on morphable mesh models \cite{Paysan09, Li17,Blanz99} can fit 3DMM shape and texture parameters to images for a given subject. However, such mesh-based approaches suffer from an inherent resolution-memory trade-off, and cannot handle topological changes caused by hair, glasses, and other accessories. 
Recent methods build on neural radiance fields~\cite{mildenhall2020} to learn personalized avatars~\cite{Gafni21, Park21b, Park21} and yield high-quality images, especially if the generated expressions are close to the training data.
A key challenge for building animatable facial avatars with implicit fields is the modeling of deformations. Previous work either achieves this by conditioning the implicit representation on expressions \cite{Gafni21} or via a separate displacement-based warping field~\cite{Park21b, Park21}. Such under-constrained formulations limit the generalization ability, requiring large numbers of training poses. 

In this paper, we propose \emph{I}mplicit \emph{M}orphable \emph{avatar} (\methodname), a novel approach for learning personalized, generalizable and 3D-consistent facial avatars from monocular videos, see Fig.~\ref{fig:teaser}.
The proposed method combines the fine-grained expression control provided by 3DMMs with the high-fidelity geometry and texture details offered by resolution-independent implicit surfaces, taking advantage of the strengths of both methods.
\methodname{}s are modeled via three continuous implicit fields, parametrized by multilayer perceptrons (MLPs), that represent the geometry, texture, and pose- and expression-related deformations. 
Inspired by FLAME~\cite{Li17}, we represent the deformation fields via \emph{learned} expression blendshapes, linear blend skinning weights, and pose correctives in the canonical space.
The blendshapes and weights are then used to warp canonical points to the deformed locations given expression and pose conditions. This pose and expression invariant formulation of shape and deformations improves generalization to unseen poses and expressions, resulting in a model of implicit facial avatars that provides fine-grained and interpretable control.

To learn from monocular videos with dynamically deforming faces, the mapping from pixels to 3D locations on the canonical surface is required. To this end, we augment the root-finding algorithm from SNARF~\cite{Chen21} with a ray marcher akin to that of IDR~\cite{Yariv20}, yielding the canonical surface correspondence for each pixel. The color prediction for each image location is then given by the canonical texture network, which formulates our primary source of supervision: a per-pixel image reconstruction objective.

To obtain gradients for the canonical point, we observe that the location of the surface-ray intersection is implicitly defined by the geometry and deformation network with two constraints:
1.~the canonical point must lie on the surface and 2.~its deformed location must be on the marched ray. Given these equality constraints, we derive an analytic gradient for the iteratively located surface intersection via implicit differentiation, which allows end-to-end training of the geometry and deformation networks from videos.

We compare our method with several hard baselines and state-of-the-art (SOTA) methods using image similarity and expression metrics. To evaluate the generated geometry under different expressions and poses, we construct a synthetic dataset containing 10 subjects. We quantitatively show that our method produces more accurate geometry and generalizes better to unseen poses and expressions. When applied to real video sequences, \methodname{} reconstructs the target poses and expressions more accurately than that state of the art (SOTA) and qualitatively achieves more accurate geometry and better extrapolation ability over poses and expressions.

\noindent In summary, we contribute:
\begin{compactitem}
\item a 3D morphing-based implicit head avatar model with detailed geometry and appearance that generalizes across diverse expressions and poses,
\item a differentiable rendering approach that enables end-to-end learning from videos, and
\item  a synthetic video dataset for evaluation.
\end{compactitem}

\begin{figure*}
  \includegraphics[width=1\textwidth]{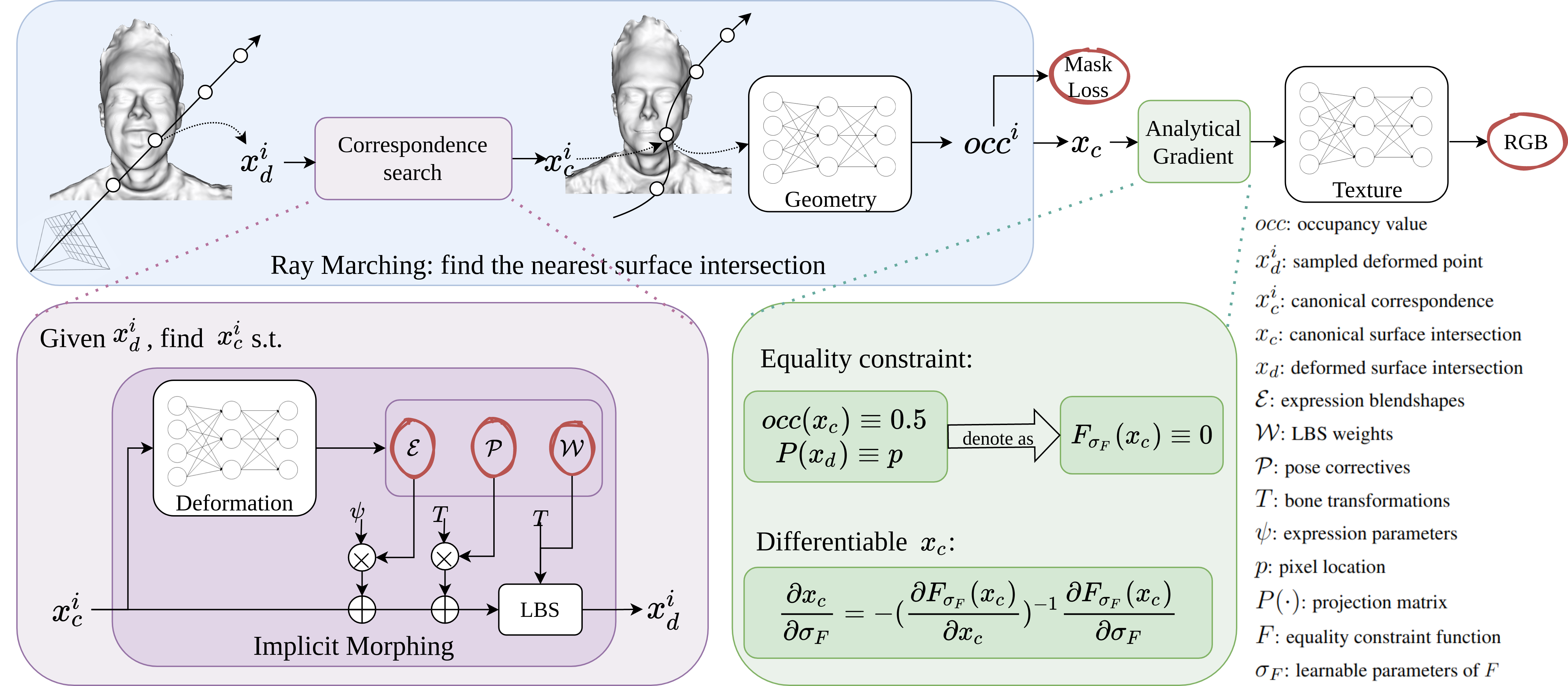}
  \caption{\textbf{Method overview.} Given a pixel location, our method performs \textbf{\textcolor{CornflowerBlue}{ray marching}} in the deformed space. For each deformed point $x_d^i$, we conduct \textbf{\textcolor{Orchid}{correspondence search}} to find the corresponding canonical point $x_c^i$. Our novel \textbf{\textcolor{Plum}{implicit morphing}} leverages the canonical blendshape and skinning-weight fields $\mathcal{E}, \mathcal{W}$ and $\mathcal{P}$ to morph the canonical point $x_c^i$ to its deformed location $x_d^i$ given expression and pose conditions. After finding the nearest canonical surface intersection $x_c$, our novel  \textbf{\textcolor{ForestGreen}{analytical gradient}} formula allows efficient computation of gradients for the geometry and deformation fields. Finally, we predict the RGB values by querying the canonical texture network. We use an image reconstruction loss and a mask loss, and optionally supervise the predicted blendshapes and skinning-weights.%
  }
  \label{fig:pipeline}
\end{figure*}

\section{Related work}
\label{sec:related}
\noindent\textbf{3D Face Models and Avatar Reconstruction.} 
Estimating 3D shape from monocular input is an ill-posed problem, traditionally addressed by using data-based statistical priors.  
The seminal work of Blanz and Vetter~\cite{Blanz99} used principal component analysis (PCA) to model facial appearance and geometry on a low-dimensional linear subspace, known as the 3D Morphable Model (3DMM).
Extensions %
include multilinear models for shape and expression~\cite{Vlasic05, Cao13}, 
full-head PCA models \cite{Dai20, Ploumpis20, Li17}, deep non-linear models~\cite{Ranjan18, Tran18} and fully articulated head models with linear blend skinning (LBS) and corrective blendshapes~\cite{Li17}. %
3DMM and its variants have been widely used within optimization-based~\cite{Gecer19, Romdhani05, Schonborn17, Thies16} and deep learning-based approaches~\cite{Deng19, Dib21, Feng21, Tewari17, Tewari18, Tewari19, Sanyal19, Shang20, Lattas20, li2018feature}. %
These methods can obtain overall accurate estimates of the facial area but typically lack details, do not model the entire head, and cannot properly represent eyes, teeth, hair, and accessories.

A related line of research estimates personalized rigs from monocular input, %
i.e.~3D representations of the head along with a set of controls that can be used for animation. 
This has been traditionally addressed
by recovering a personalized set of blendshape bases, %
obtained through deformation transfer~\cite{Sumner04, Garrido16, Ichim15, Cao16, Hu17} or deep neural networks~\cite{Bai21, Chaudhuri20, Yang20}. 
To represent detailed geometry, a few methods additionally reconstruct a layer of mid- or fine-level correctives that can be deformed along with the underlying coarse mesh~\cite{Feng21, Ichim15, Garrido16, Yang20}. 
Here, we present a new approach that can recover a higher-fidelity facial rig than prior work, and is controlled by expression blendshapes as well as jaw, neck, and eye pose parameters.

\smallskip
\noindent\textbf{Neural Face Models.}
The recent success of neural implicit shape representations~\cite{Chen19, Kellnhofer21, Mescheder19, Mildenhall20, Park19, Yariv20} has led to several methods that build 3D facial models within this paradigm. 
Yenamandra \etal~\cite{Yenamandra21} propose an implicit morphable model  that decouples shape, expression, appearance, and hair style. 
Ramon~\etal~\cite{Ramon21} estimate a full head model from a few input images by pre-training signed distance fields on a large number of raw 3D scans. 
These works demonstrate an improved ability to represent the full head and torso along with detailed geometry, %
but the estimated shapes cannot be animated. In contrast, our implicit surface representation can be controlled via 3DMM parameters.%

Neural volumetric representations such as NeRF~\cite{Mildenhall20} have also been explored in the context of faces~\cite{Gafni21, Chan21, Wang21}. These representations enable the encoding of thin structures such as hair, and they can model complex surface/light interactions.  
Wang \etal~\cite{Wang21} propose a localized compositional model that combines discrete low-resolution voxels with neural radiance fields. Their method requires a complex multi-view video system for training.
The closest work to ours is NerFACE~\cite{Gafni21}, which recovers an animatable neural radiance field from a single monocular video by conditioning on the expression parameters from a 3DMM. While NerFACE achieves good image quality on novel views and interpolated expressions, it struggles to extrapolate to unseen expressions, 
and the quality of the geometry is too noisy to be used in 3D settings (Fig.~\ref{fig:real_qualitative}).

\smallskip
\noindent\textbf{Implicit deformation fields.} 
Modeling dynamic objects with implicit neural fields is an active research topic, with currently three main approaches. 
The first is to condition each frame on a latent code, 
\eg a time stamp \cite{Xian21}, a learned latent vector \cite{Park21, Wang21}, or a vector from a pre-computed parametric model \cite{Deng20, Gafni21, Mihajlovic21, Saito21}. 
A second, possibly complementary approach, is to use a ``backward'' deformation field. This is an additional neural network that maps observations in deformed space to observations in canonical space, where the implicit function is then evaluated \cite{Niemeyer19, Park21, Park21, Pumarola21, Tiwari21, Tretschk21, Saito21}. The deformation fields are modeled as velocity fields~\cite{Niemeyer19}, translation fields~\cite{Pumarola21, Tiwari21}, rigid transformations~\cite{Park21}, or skinning-weight fields~\cite{Jeruzalski20, Mihajlovic21, Saito21}. 
While they have demonstrated an impressive capacity for learning correspondences even in the absence of ground-truth supervision, the backwards formulation makes them pose-dependent and hence requires a large dataset for learning, showing reduced generalization when deformations are too far away from the training set. 
To tackle this, \emph{forward} deformation fields have been recently proposed \cite{Chen21} for learning 3D body avatars. These learn a continuous forward skinning weight field, and corresponding canonical points are found using iterative root finding. Concurrent work extends the idea to model textured \cite{Dong2022pina, Jiang22} and generative human avatars \cite{Chen2022gdna}. To improve generalization outside the scope of train-time expressions, here we extend the idea of forward skinning to the problem of facial deformations and propose a new analytical gradient formula that allows the deformation fields to be directly learned from videos.

\section{Method}
We propose \methodname{}, an implicit morphable head avatar that equips implicit surfaces with fine-grained expression control by leveraging morphing-based deformation fields.
In this section, we first recap the deformation formulation of the FLAME face model~\cite{Li17}, followed by the representations for the canonical geometry, deformation, and texture fields. Then, we introduce correspondence search to find canonical points for image pixels and derive the analytical gradients for end-to-end training.

\subsection{Recap: FLAME Face Morphable Model}
The FLAME face model~\cite{Li17} parameterizes facial geometry with shape, pose, and expression components. Since we focus on personal facial avatars, we specifically represent the pose- and expression-dependent shape variations. The simplified FLAME mesh model is denoted by: 
\begin{equation}
   M(\boldsymbol{\theta}, \boldsymbol{\psi}) = LBS(T_P(\boldsymbol{\theta}, \boldsymbol{\psi}), J(\boldsymbol{\psi}), \boldsymbol{\theta}, \mathcal{W}), 
\end{equation}
where $\boldsymbol{\theta}$ and $\boldsymbol{\psi}$ denote the pose and expression parameters, and $LBS(\cdot)$ and $J(\cdot)$ define the standard skinning function and the joint regressor, respectively. $\mathcal{W}$ represents the per-vertex skinning weights for smooth blending, and $T_P$ denotes the canonical vertices after adding expression and pose correctives, represented as:
\begin{equation}
   T_P(\boldsymbol{\theta}, \boldsymbol{\psi}) = \boldsymbol{\bar{T}} + B_E(\boldsymbol{\psi}; \mathcal{E}) + B_P(\boldsymbol{\theta}; \mathcal{P}), 
\end{equation}
where $\boldsymbol{\bar{T}}$ is the personalized canonical template. $B_P(\cdot)$ and $B_E(\cdot)$ calculate the additive pose and expression offsets using the corrective blendshape bases $\mathcal{P}$ and $\mathcal{E}$ given the animation conditions $\theta$ and $\psi$.
Our method extends the discrete $\mathcal{W}$, $\mathcal{E}$, and $\mathcal{P}$ defined on vertices to be continuous fields represented by MLPs, making it possible to morph continuous canonical representations.

\subsection{\methodname}
\methodname is represented by three neural implicit fields, defining the canonical geometry, deformation bases, and texture of the person, as shown in Fig.~\ref{fig:pipeline}. Details of the network architecture can be found in the \suppmat.

\smallskip
\noindent\textbf{Geometry.}
We represent the canonical geometry using an MLP that predicts the occupancy values for each canonical 3D point. We additionally condition the geometry network $\function{f}$ on a per-frame learnable latent code $\boldsymbol{l} \in \mathbb{R}^{n_l}$, similar to NerFace \cite{Gafni21}, and leverage positional encoding \cite{Mildenhall20} to encourage high frequency details in the canonical geometry
\begin{equation}
\function{f}(x, l): \mathbb{R}^3 \times  \mathbb{R}^{n_l} \rightarrow occ .
\end{equation}

\smallskip
\noindent\textbf{Deformation.}
Following FLAME \cite{Li17}, our deformation network $\function{d}$ predicts the additive expression blendshape vectors $\mathcal{E} \in \mathbb{R}^{n_{e} \times 3}$, the pose correctives $\mathcal{P} \in \mathbb{R}^{n_{j} \times 9 \times 3}$, and the linear blend skinning weights $\mathcal{W} \in \mathbb{R}^{n_{j}}$ for each point in the canonical space, where $n_e$ and $n_j$ denote the number of expression parameters and bone transformations
\begin{equation}
    \function{d}(x): \mathbb{R}^3 \rightarrow \mathcal{E, P, W}.
\end{equation}
In a slight abuse of notation we reuse $\mathcal{E}$, $\mathcal{P}$, and $\mathcal{W}$ from FLAME -- please note that these denote continuous implicit fields from here on. 
For each canonical point $x_c$, the transformed location $x_d\coloneqq w_{\sigma_d}(x_c) $ is:
\begin{equation}
    x_d = LBS(x_c + B_P(\boldsymbol{\theta}; \mathcal{P}) + B_E(\boldsymbol{\psi}; \mathcal{E}), J(\boldsymbol{\psi}), \boldsymbol{\theta}, \mathcal{W}),
\end{equation}
where $J$ is the joint regressor from FLAME. This defines the forward mapping from canonical points $x_c$ to deformed locations $x_d$. We detail the computation of the inverse mapping from deformed to canonical space in Sec.~\ref{sec:diff_rendering}.

\smallskip
\noindent\textbf{Normal-conditioned texture.} 
We leverage a texture MLP $\function{c}$ to map each location in the canonical space to an RGB color value. To account for non-uniform lighting effects, we additionally condition the texture network on the normal direction of the deformed shape. For implicit surfaces, the normal direction can be calculated as the normalized gradient of the occupancy field w.r.t.~the 3D location. In our case, the gradient of the deformed shape is given by:
\begin{equation}
    \pardev{\function{f}(x_c)}{x_d} = \pardev{\function{f}(x_c)}{x_c}  \pardev{x_c}{x_d} = \pardev{\function{f}(x_c)}{x_c} \left(\pardev{w_{\sigma_d}(x_c)}{x_c}\right)^{-1}.
\end{equation}
Since the appearance in the mouth region cannot be modeled purely by warping due to dis-occlusions \cite{Park21}, our final predicted color $c$ is calculated from the canonical location $x_c$, the normal direction of the deformed shape $n_d$, and the jaw pose and expression parameters $\boldsymbol{\theta}$ and $\boldsymbol{\psi}$
\begin{equation}
    \function{c}(x_c, n_d, \boldsymbol{\theta}, \boldsymbol{\psi}):  \mathbb{R}^3 \times \mathbb{R}^3 \times \mathbb{R}^3 \times \mathbb{R}^{50} \rightarrow c.
\end{equation}

\subsection{Differentiable Rendering}
\label{sec:diff_rendering}
To optimize the canonical networks from videos with expressions and poses, we first introduce non-rigid ray marching to find the canonical surface point for each ray, and introduce analytical gradients that enable end-to-end training of the geometry and deformation networks.

\smallskip
\noindent\textbf{Non-rigid ray marching.}
Given a camera location $r_o$ and a ray direction $r_d$ in the deformed space, we follow IDR \cite{Yariv20} and perform ray marching in the deformed space.
To determine the occupancy values for the sampled points $x_d$, we follow SNARF~\cite{Chen21} and leverage iterative root finding to locate the canonical correspondences $x_c$ and query their occupancy values. Thus, we can locate the nearest canonical surface intersection for each ray iteratively.

\smallskip
\noindent\textbf{Gradient.}
To avoid back-propagation through the iterative process, we derive analytical gradients for the location of the canonical surface point $x_c$, leveraging that $x_c$ must satisfy the surface and ray constraints: %
\begin{eqnarray}
    \function{f}(x_c) & \equiv  & 0.5, \\ 
    (w_{\sigma_d}(x_c) - r_o)\times r_d & \equiv & 0, 
\end{eqnarray}
where 0.5 is defined as the level set for the surface. For convenience, we rewrite this equality constraint as $F_{\sigma_F}(x_c)\equiv 0$,
which implicitly defines the canonical surface intersection $x_c$. The learnable parameters of the geometry and deformation networks are  $\sigma_F = \sigma_f \cup \sigma_d$. 
We leverage implicit differentiation to attain the gradient of $x_c$ w.r.t. the parameters of the geometry and deformation networks:
\begin{equation}
\begin{aligned}
    &\frac{dF_{\sigma_F}(x_c)}{d\sigma_F} = 0\\
    \Leftrightarrow ~& \frac{\partial F_{\sigma_F}(x_c)}{\partial \sigma_F} + \frac{\partial F_{\sigma_F}(x_c)}{\partial x_c}\frac{\partial x_c}{\partial \sigma_F} = 0\\
    \Leftrightarrow ~& \frac{\partial x_c}{\partial \sigma_F} = -(\frac{\partial F_{\sigma_F}(x_c)}{\partial x_c})^{-1}\frac{\partial F_{\sigma_F}(x_c)}{\partial \sigma_F}.
\end{aligned}
\end{equation}
We also supervise non-surface rays with a mask loss (Eq.~\ref{Eq:mask_loss}), in which case the equality constraint is defined as $w_{\sigma_d}(x_c)\equiv x_d^{\star}$, where $x_d^{\star}$ is the point on the ray with the smallest occupancy value, located via ray sampling. %

\subsection{Training Objectives}
\label{sec:objectives}
\textbf{RGB loss} supervises the rendered pixel color:
\begin{equation}
    \loss_{RGB} = \frac{1}{|P|}\sum_{p \in P^{in}}\|C_p - \function{c}(x_c)\|_1,
\end{equation}
where $P$ denotes the set of training pixels, and $P^{in} \subset P$ denotes the foreground pixels where a ray-intersection has been found. $C_p$ and $\function{c}(x_c)$ represent the ground-truth and predicted RGB values of pixel $p$. The analytical gradient formula enables $\loss_{RGB}$ to not only optimize texture, but also the geometry and deformation networks.

\textbf{Mask loss} supervises the occupancy values for non-surface rays $p \in  P \setminus P^{in}$:
\begin{equation}
    \loss_{M} = \frac{1}{|P|}\sum_{p \in  P \setminus P^{in}}CE(O_p, \function{f}(x_c^{\star})),
    \label{Eq:mask_loss}
\end{equation}
where $CE(\cdot)$ is the cross-entropy loss calculated between the ground-truth $O_p$ and predicted occupancy values $\function{f}(x_c^{\star})$. The mask loss also optimizes the deformation network thanks to the analytical gradient.

An \textit{optional} \textbf{FLAME loss} leverages prior knowledge about expression and pose deformations from FLAME~\cite{Li17} by supervising the deformation network with the corresponding values of the nearest FLAME vertices:
\begin{eqnarray}
\lefteqn{\loss_{FL} = \frac{1}{|P|}\sum_{p \in P^{in}}(\lambda_e\|\mathcal{E}_p^{GT} - \mathcal{E}_p\|_2} \nonumber \\
   & &  +\lambda_p \|\mathcal{P}_p^{GT} - \mathcal{P}_p\|_2 + \lambda_w\|\mathcal{W}_p^{GT} - \mathcal{W}_p\|_2),
     \label{Eq:flame_loss}
\end{eqnarray}
where $\mathcal{E}_p$, $\mathcal{P}_p$, and $\mathcal{W}_p$ denote the predicted values of the deformation network, and $\mathcal{E}_p^{GT}$, $\mathcal{P}_p^{GT}$ and $\mathcal{W}_p^{GT}$ denote the pseudo ground truth defined by the nearest FLAME vertices. We set $\lambda_e=\lambda_p=1000$, and $\lambda_w=0.1$ for our experiments.
Our final training loss is
\begin{equation}
    \loss_{} = \loss_{RGB} + \lambda_{M}\loss_{M} + \lambda_{FL}\loss_{FL},
\end{equation}
where $\lambda_{M}=2$ and $\lambda_{FL}=1$.
\section{Experiments}
\begin{figure}[t]
\begin{center}
\setlength\tabcolsep{1pt}
\newcommand{\crop}{0.8cm}
\newcommand{\cropsmall}{0.4cm}
\newcommand{\height}{1.5cm}

\begin{tabularx}{\textwidth}{lccccccccccc}
\rotatebox[origin=l]{90}{D-Net} & \includegraphics[height=\height]{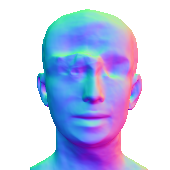} & \includegraphics[height=\height]{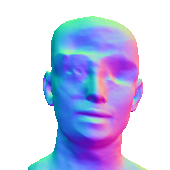} & \includegraphics[height=\height]{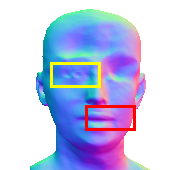} & \includegraphics[height=\height]{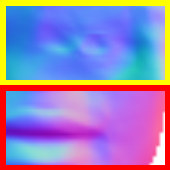} & \includegraphics[height=\height]{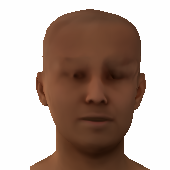} & \\
\rotatebox[origin=l]{90}{B-Morph} & \includegraphics[height=\height]{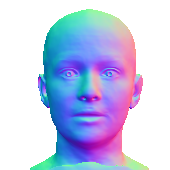} & \includegraphics[height=\height]{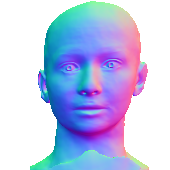} & \includegraphics[height=\height]{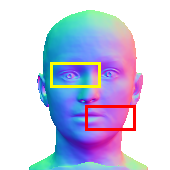} & \includegraphics[height=\height]{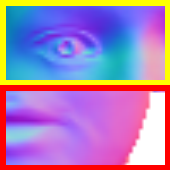} & \includegraphics[height=\height]{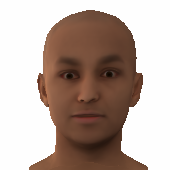} & \\
\rotatebox[origin=l]{90}{C-Net} & \includegraphics[height=\height]{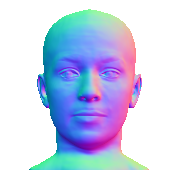} & \includegraphics[height=\height]{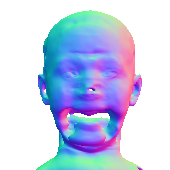} & \includegraphics[height=\height]{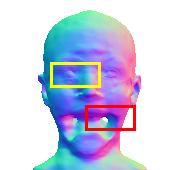} & \includegraphics[height=\height]{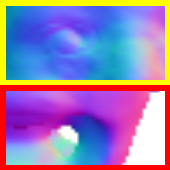} & \includegraphics[height=\height]{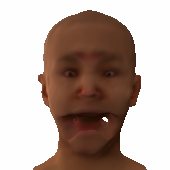} & \\
\rotatebox[origin=l]{90}{Fwd-Skin} & \includegraphics[height=\height]{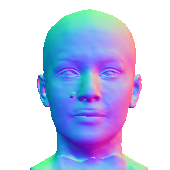} & \includegraphics[height=\height]{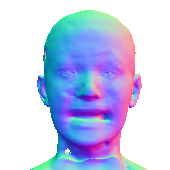} & \includegraphics[height=\height]{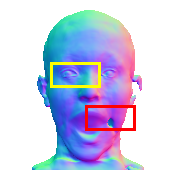} & \includegraphics[height=\height]{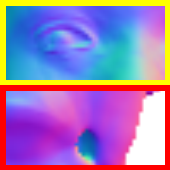} & \includegraphics[height=\height]{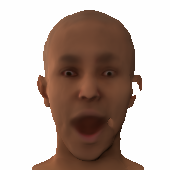} & \\
\rotatebox[origin=l]{90}{\textbf{Ours}} & \includegraphics[height=\height]{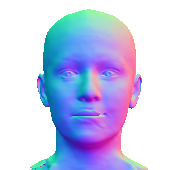} & \includegraphics[height=\height]{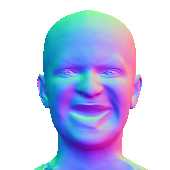} & \includegraphics[height=\height]{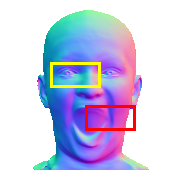} & \includegraphics[height=\height]{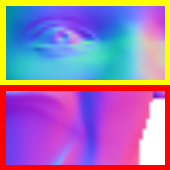} & \includegraphics[height=\height]{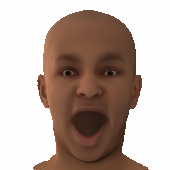} & \\
\rotatebox[origin=l]{90}{GT} & \includegraphics[height=\height]{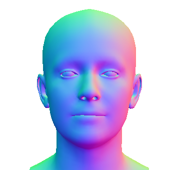} & \includegraphics[height=\height]{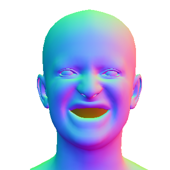} & \includegraphics[height=\height]{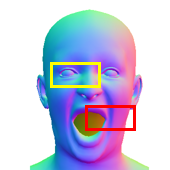} & \includegraphics[height=\height]{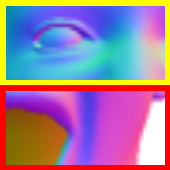} & \includegraphics[height=\height]{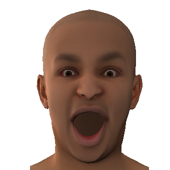} & \\
& Neutral & Medium & Strong & Zoom & RGB \\
\end{tabularx}
\caption{
\textbf{Qualitative results on synthetic data.}  
As the expression strength increases from left to right, baseline methods either collapse to a neutral expression (D-Net, B-Morph) or produce invalid geometry (C-Net, Fwd-Skin). In contrast, our method successfully handles even the most extreme expressions.
}

\label{fig:synthetic_qualitative}
\end{center}
\end{figure}

\label{sec:results}
This section empirically evaluates the benefits of the proposed approach in terms of geometry accuracy and expression generalization. We conduct experiments on both synthetic data with known geometry and real video sequences.

\mcb{Our approach improves geometry by 4\% compared with the best performing baseline (Tab.~\ref{tab:ablations}) and reproduces expressions 13\% more faithfully than NerFACE~\cite{Gafni21} (Tab.~\ref{tab:sota}) while being on par in terms of perceptual quality.} \yf{it's rather painful that we compare geometry with the best performing baseline which we made up ourselves and has never been proposed in previous works.}
\subsection{Datasets}

\noindent
\textbf{Synthetic Dataset.} 
We conduct controlled experiments on a synthetic dataset by rendering posed and textured FLAME meshes. 
For the \emph{training} set, we render a video that is representative of a speech sequence. We take FLAME expression parameters from the VOCA speech dataset \cite{Cudeiro19} and head poses fitted from real videos.  %
We build the \emph{test} set 
with unseen, stronger expressions extracted from COMA \cite{Ranjan18}.\mcb{Some of them might have been seen. Why not write "...test set that also includes stronger expressions such as big smiles.}
Our synthetic dataset consists of 10 subjects with varied facial shapes and appearance, with an average of 5,368 frames for training and 1876 frames for testing per subject. For testing, we subsample every 10th frame. 
We release the synthetic dataset for research purposes.

\smallskip
\noindent
\textbf{Real Video Dataset.}
We evaluate on real videos from a single stationary camera. We calculate foreground masks with MODNet \cite{Ke20} and estimate the initial FLAME parameters using DECA \cite{Feng21}, which are refined by fitting to 2D facial keypoints \cite{Bulat17}. Please see Sup.~Mat.~for more details.
The real video dataset consists of 4 subjects, with roughly 4,000 frames for training and 1,000 frames for testing per subject. The \emph{training} videos cover mostly neutral expressions in a speech video, while the \emph{test} videos include unseen, difficult expressions such as jaw opening, big smiles, and more. We subsample every 10th frame for testing.

\subsection{Ablation Baselines}
This paper tackles the key difficulty in building animatable avatars: capturing the per-frame deformations with respect to the canonical shape.
We compare our method with the commonly used previous approaches by replacing our deformation module with the following alternatives:

\smallskip
\noindent
\textbf{Pose- and expression-conditioned network (C-Net).}
C-Net is inspired by NerFACE~\cite{Gafni21} but is designed for implicit surfaces. It first applies a rigid transformation to the deformed shape, which brings the full upper body to the canonical space with the inverse head pose transformation; it then models other deformations by conditioning on the pose and expression parameters.

\smallskip
\noindent
\textbf{Displacement warping (D-Net).}
D-Net uses a deformation network with pose and expression parameters as input, and predicts displacement vectors for deformed points,  warping them to the canonical space. The predicted displacements are supervised with FLAME, similar to Eq.~\ref{Eq:flame_loss}.

\smallskip
\noindent
\textbf{Backward morphing (B-Morph).}
B-Morph leverages the morphing formulation of FLAME and predicts expression blendshapes, pose corrective vectors, and LBS weights. However, the deformation network is conditioned on the deformed location as well as pose and expression parameters and performs backward morphing. %
In contrast, our deformation network takes as input only the canonical point, which is pose- and expression-independent, enabling better generalization~\cite{Chen21}. The learned blendshapes and weights of this baseline are supervised by FLAME pseudo GT.

\smallskip
\noindent
\textbf{Forward skinning + expression-conditioning (Fwd-Skin).}
This baseline is adapted from SNARF \cite{Chen21}, which was originally proposed for human body avatars. Here, the deformation network only models the LBS weights, while the expression and pose-related correctives are handled by conditioning on the geometry and texture networks.

\smallskip
\noindent
\textbf{IMavatar unsupervised (Ours-).} This baseline eliminates the FLAME pseudo GT supervision, learning solely from images and masks (only used for exp.~with real data). 
\subsection{Metrics}
\label{sec:metrics}
The goal of this work is to obtain an animatable 3D head from a video, and hence we evaluate the geometric accuracy (only available for the synthetic dataset), image quality, and expression fidelity.
Image quality is measured via the Manhattan distance ($L_1$), SSIM, PSNR and LPIPS \cite{Zhang18} metrics, following the practice in NerFACE~\cite{Gafni21}. 
To measure geometric consistency for synthetic data, we report the average angular normal error between the generated normal map and ground truth, denoted as the \emph{Normals} metric in Tab.~\ref{tab:ablations} \oh{I think this is typically called Normal Consistency (NC). Chamfer distance would be a better metric for geometry - why do we not use that}. Since we focus on modeling deformation-related geometry and texture, both normal consistency and image similarity metrics are measured in the face interior region.
For both synthetic and real data, we measure the expression fidelity by calculating the distance between generated and (pseudo) GT facial keypoints. 
We estimate the facial keypoints of predicted images with~\cite{Bulat17}, and (pseudo) GT keypoints are obtained from posed FLAME meshes.
\va{Somewhere we should say that predicted images means that we input known expression and pose and compare against ground-truth}
\subsection{Results on Synthetic Dataset}
\begin{table}
\centering
\resizebox{\linewidth}{!}{ %
\small
\begin{tabular}{l|ccccccc}
\toprule
Method & Expression $\downarrow$ & Normals $\downarrow$& $L_1$ $\downarrow$ & PSNR $\uparrow$ & SSIM $\uparrow$ & LPIPS $\downarrow$ \\ %
\midrule
C-Net & 3.248 & 9.108 & 0.02245 & 26.67 & 0.9829 & 0.03053\\
D-Net & 7.452 & 26.174 & 0.07881 & 19.62 & 0.9481 & 0.05881\\
B-Morph & 4.941 & 12.150 & 0.03293 & 24.95 & 0.9726 & 0.03340\\
Fwd-Skin & 2.825 & 8.130 & 0.01920 & 27.30 & 0.9852 & 0.02812\\
Ours & \textbf{2.558} & \textbf{5.901} & \textbf{0.01807} & \textbf{28.75} & \textbf{0.9900} & \textbf{0.01581}\\
\bottomrule
\end{tabular}
} %
\caption{
\textbf{Quantitative results for synthetic experiment.} Compared to the baselines, our method achieves more consistent surface normals, better image quality, and more accurate expressions.%
} %
\label{tab:ablations}
\end{table}

\begin{figure}[t]
\centerline{
\includegraphics[width=\linewidth]{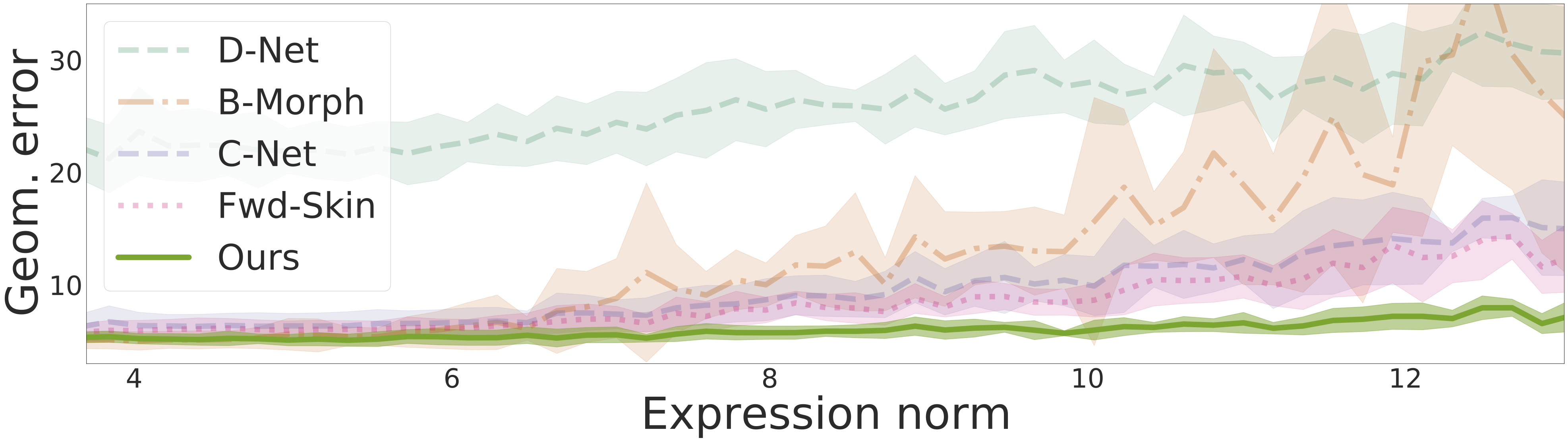}
}
\caption{
\textbf{Expression extrapolation.} Performance of baseline methods worsen drastically as expressions become more extreme (higher norm). 
\emph{Geom.~error} denotes the angular error of the surface normals (lower is better, see Sec.~\ref{sec:metrics}).
}
\label{fig:plot_synthetic_examples}
\end{figure}
We train IMavatar and baseline methods for the 10 synthetic identities and measure geometry, expression and image reconstruction errors on 12 sequences with renderings from the COMA dataset. We outperform all baselines by a large margin on all metrics (Tab.~\ref{tab:ablations}).

\paragraph{Extrapolation}
While other methods are limited to interpolation, our method is capable of extrapolating beyond seen expressions and poses. In Fig.~\ref{fig:plot_synthetic_examples}, we plot the geometric error for different strength of expressions. Most methods perform well for mild expressions (small expression norm). For stronger expressions, however, their errors increase significantly. In contrast, our method only incurs a slight increase even for strong expressions (large norm). See Sup.~Mat.~for an analogous plot for the jaw pose.
Figure \ref{fig:synthetic_qualitative} shows visual examples for neutral, medium, and strong expressions.

\subsection{Result on Real Video sequences}
\begin{figure}
\begin{center}
\setlength\tabcolsep{1pt}
\newcommand{\crop}{0.8cm}
\newcommand{\cropsmall}{0.4cm}
\newcommand{\height}{1.4cm}
\begin{tabularx}{\textwidth}{l ccccc}
\rotatebox{90} {~~~Exp~~~}
&\includegraphics[height=\height]{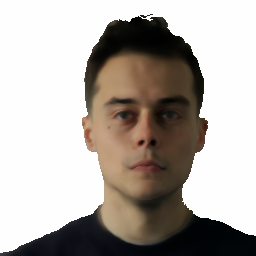}
&\includegraphics[height=\height]{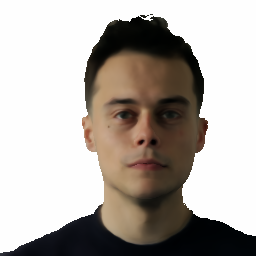}
&\includegraphics[height=\height]{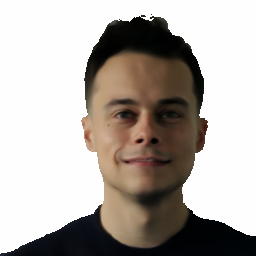}
&\includegraphics[height=\height]{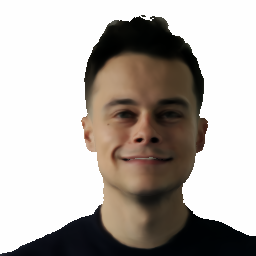}
&\includegraphics[height=\height]{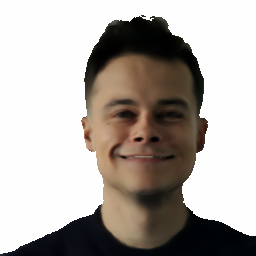}
\\
&\multicolumn{5}{c}{\includegraphics[width=0.9\linewidth]{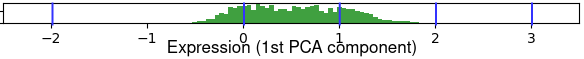}}\\
\rotatebox{90} {~~~Jaw~~~}
&\includegraphics[height=\height]{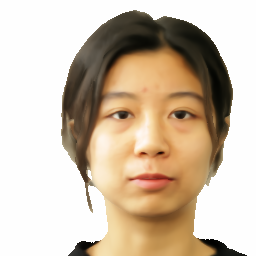}
&\includegraphics[height=\height]{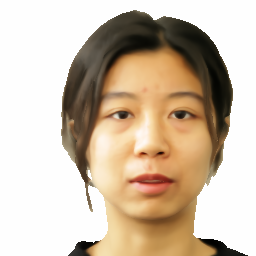}
&\includegraphics[height=\height]{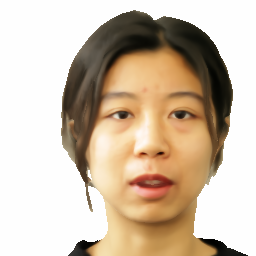}
&\includegraphics[height=\height]{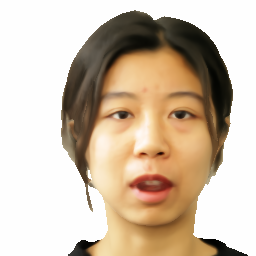}
&\includegraphics[height=\height]{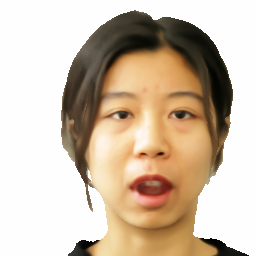}
\\
&\multicolumn{5}{c}{\includegraphics[width=0.9\linewidth]{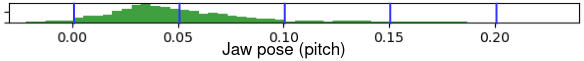}}\\
\rotatebox{90} {~~~Neck~~~}
&\includegraphics[height=\height]{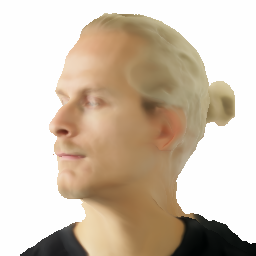}
&\includegraphics[height=\height]{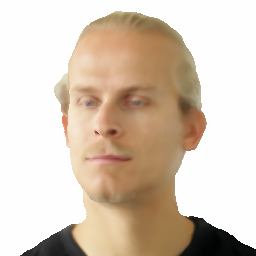}
&\includegraphics[height=\height]{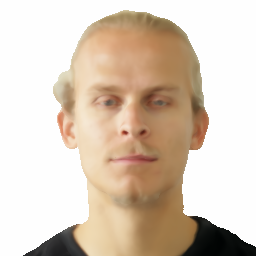}
&\includegraphics[height=\height]{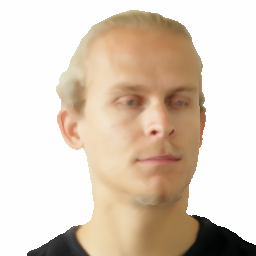}
&\includegraphics[height=\height]{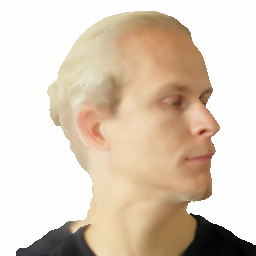}
\\
&\multicolumn{5}{c}{\includegraphics[width=0.9\linewidth]{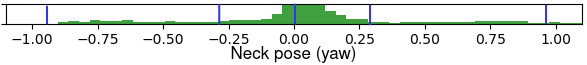}}

\end{tabularx}
\vspace{-0.1in}
\caption{
\textbf{Expression and pose control beyond the training distribution.}
We show interpolation and extrapolation results along with histograms of the training distribution, where the blue bars correspond to the 5 visualized samples. Our method  generalizes to expression and pose conditions outside the training distribution. %
\label{fig:real_interpolate_extrapolate}
}
\end{center}

\end{figure}
\begin{figure*}
\begin{minipage}{443pt}
\begin{center}
\setlength\tabcolsep{1pt}
\newcommand{\crop}{0.8cm}
\newcommand{\cropsmall}{0.4cm}
\newcommand{\height}{2.cm}
\begin{tabularx}{\textwidth}{ccccccccc}
D-Net & B-Morph & Fwd-Skin & C-Net & NerFACE~\cite{Gafni21} & Ours- & \textbf{Ours} & GT &
\multirow{8}{*}{\rotatebox[origin=l]{-90}{$\underrightarrow{\text{\hspace{17em}Expression and pose extrapolation\hspace{17em}}}$}}
\\
\includegraphics[height=\height]{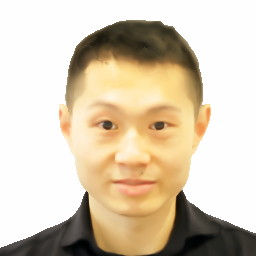}
&\includegraphics[height=\height]{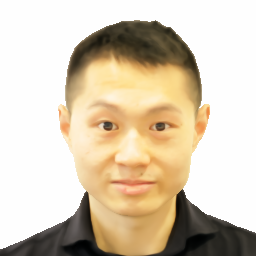}
&\includegraphics[height=\height]{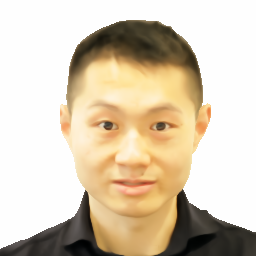}
&\includegraphics[height=\height]{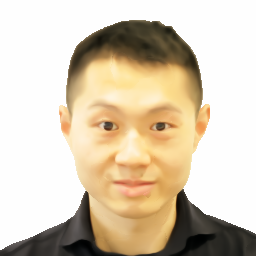}
&\includegraphics[height=\height]{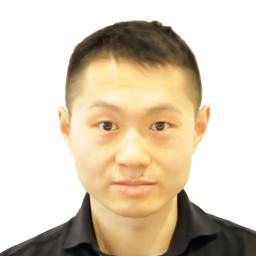}
&\includegraphics[height=\height]{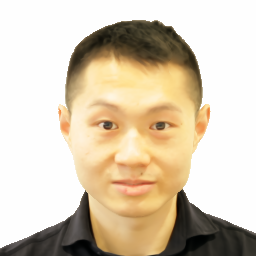}
&\includegraphics[height=\height]{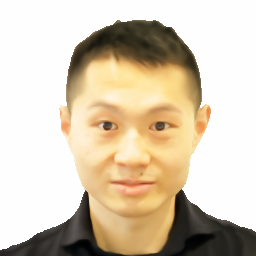}
&\includegraphics[height=\height]{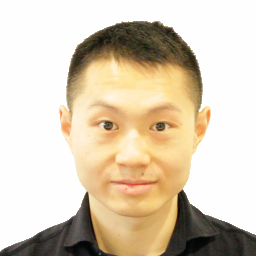}
& 
\\
\includegraphics[height=\height]{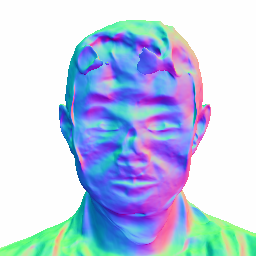}
&\includegraphics[height=\height]{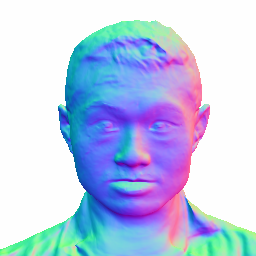}
&\includegraphics[height=\height]{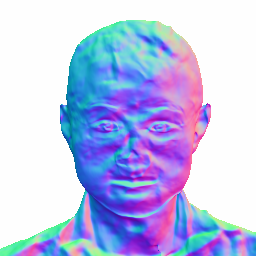}
&\includegraphics[height=\height]{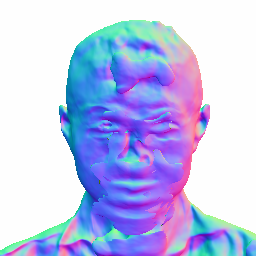}
&\includegraphics[height=\height]{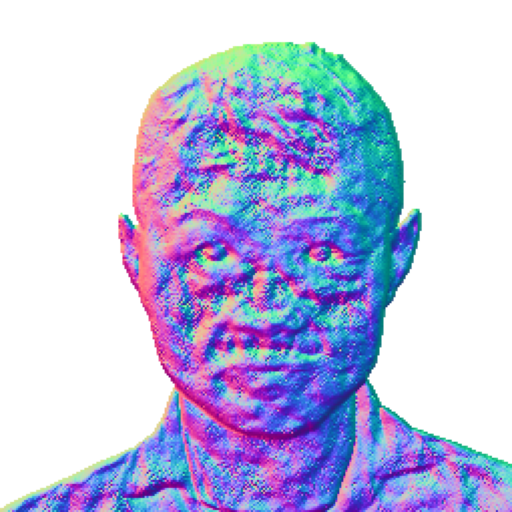}
&\includegraphics[height=\height]{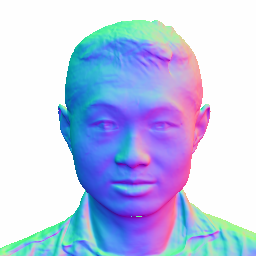}
&\includegraphics[height=\height]{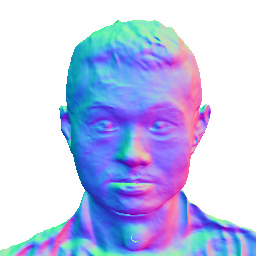}
& 
\\
\includegraphics[height=\height]{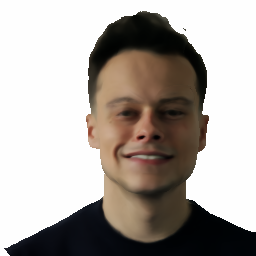}
&\includegraphics[height=\height]{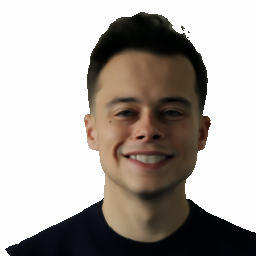}
&\includegraphics[height=\height]{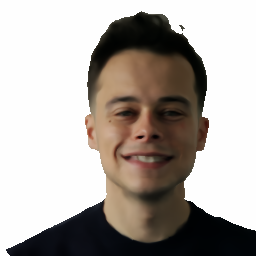}
&\includegraphics[height=\height]{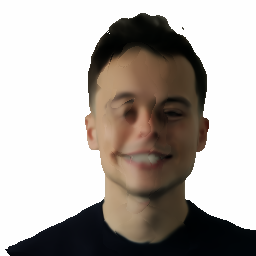}
&\includegraphics[height=\height]{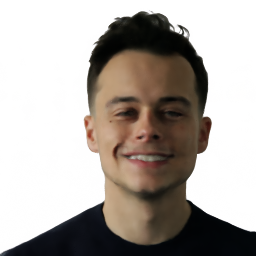}
&\includegraphics[height=\height]{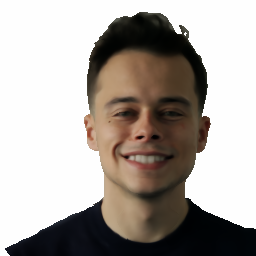}
&\includegraphics[height=\height]{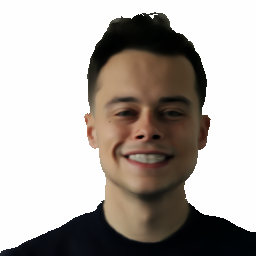}
&\includegraphics[height=\height]{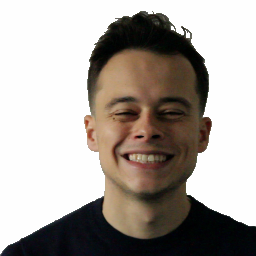}
\\
\includegraphics[height=\height]{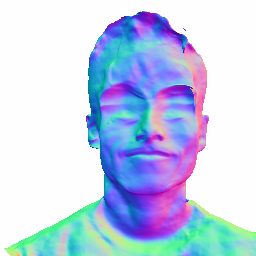}
&\includegraphics[height=\height]{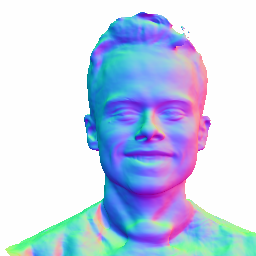}
&\includegraphics[height=\height]{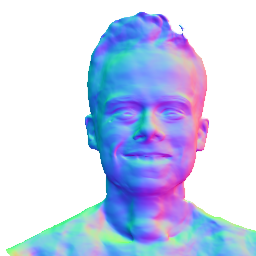}
&\includegraphics[height=\height]{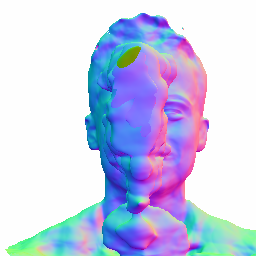}
&\includegraphics[height=\height]{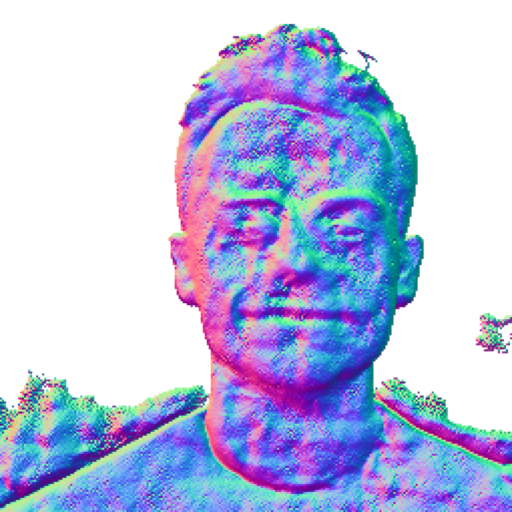}
&\includegraphics[height=\height]{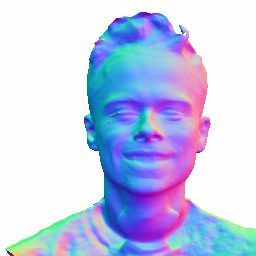}
&\includegraphics[height=\height]{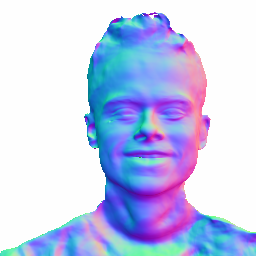}
& 
\\
\includegraphics[height=\height]{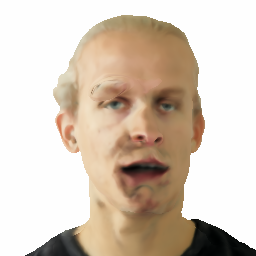}
&\includegraphics[height=\height]{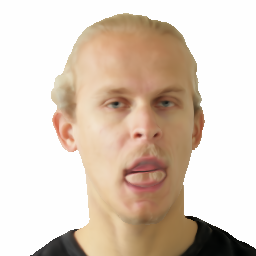}
&\includegraphics[height=\height]{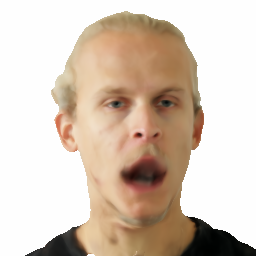}
&\includegraphics[height=\height]{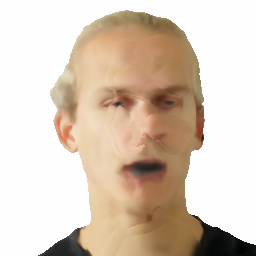}
&\includegraphics[height=\height]{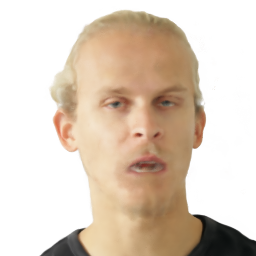}
&\includegraphics[height=\height]{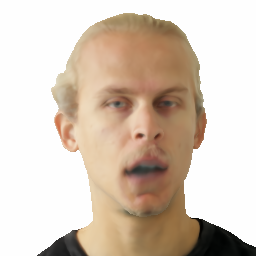}
&\includegraphics[height=\height]{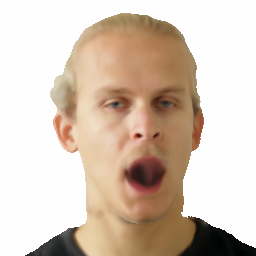}
&\includegraphics[height=\height]{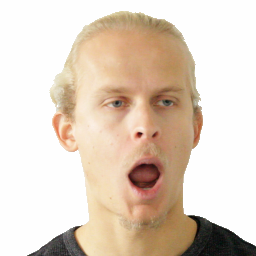}
\\
\includegraphics[height=\height]{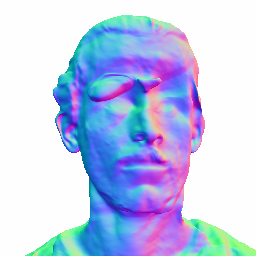}
&\includegraphics[height=\height]{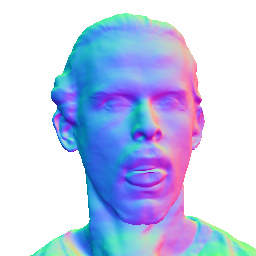}
&\includegraphics[height=\height]{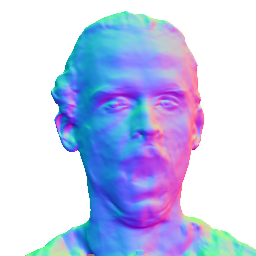}
&\includegraphics[height=\height]{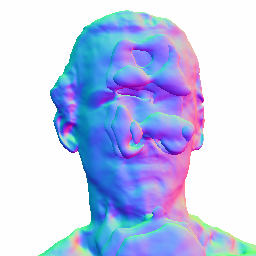}
&\includegraphics[height=\height]{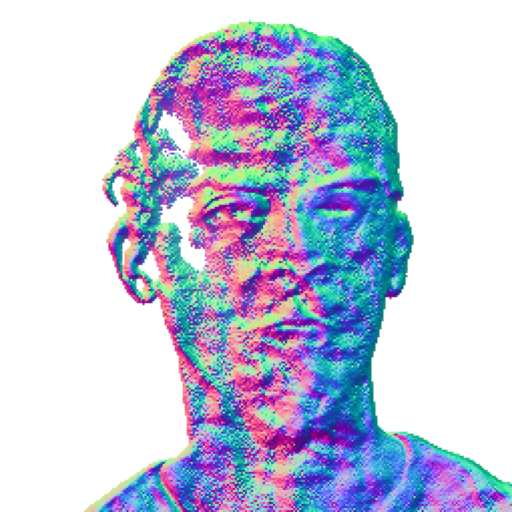}
&\includegraphics[height=\height]{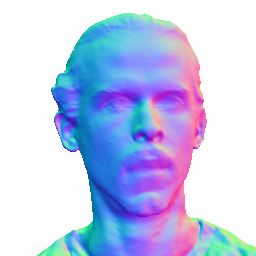}
&\includegraphics[height=\height]{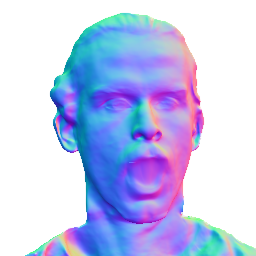}
& 
\\
\includegraphics[height=\height]{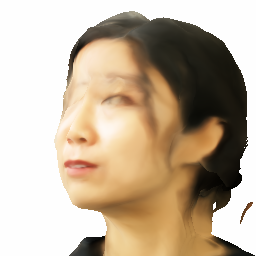}
&\includegraphics[height=\height]{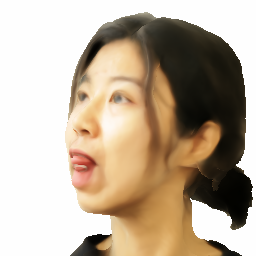}
&\includegraphics[height=\height]{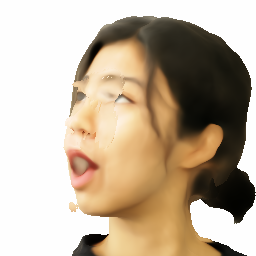}
&\includegraphics[height=\height]{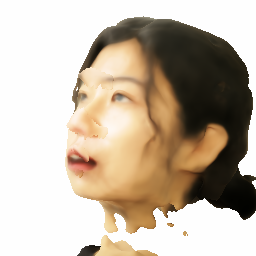}
&\includegraphics[height=\height]{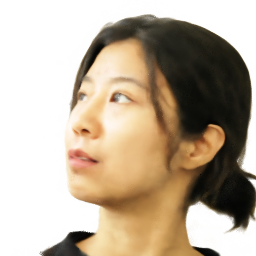}
&\includegraphics[height=\height]{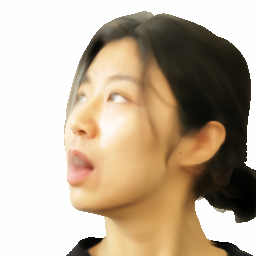}
&\includegraphics[height=\height]{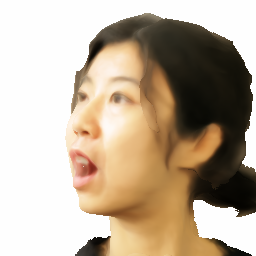}
&\includegraphics[height=\height]{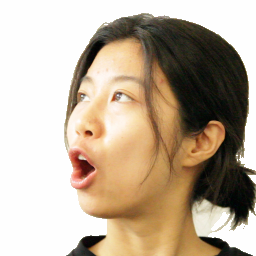}
\\
\includegraphics[height=\height]{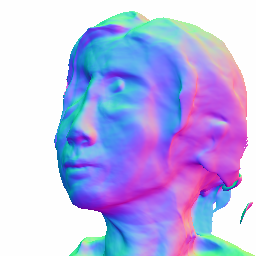}
&\includegraphics[height=\height]{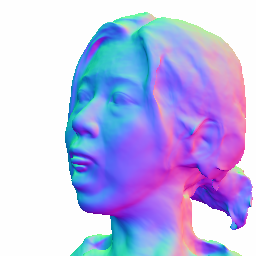}
&\includegraphics[height=\height]{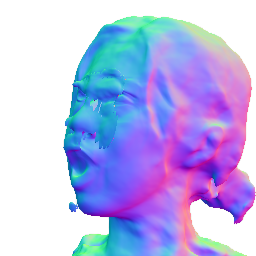}
&\includegraphics[height=\height]{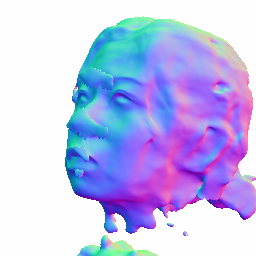}
&\includegraphics[height=\height]{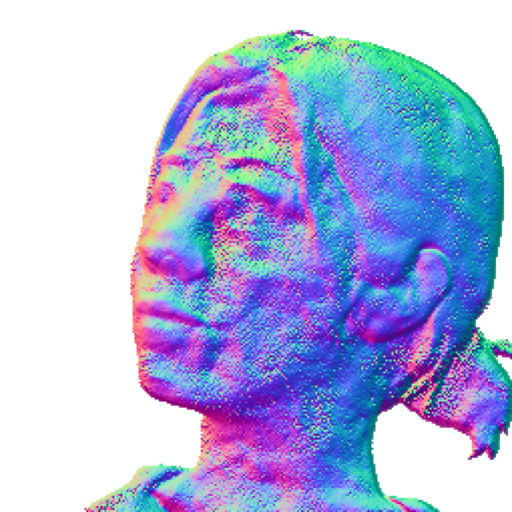}
&\includegraphics[height=\height]{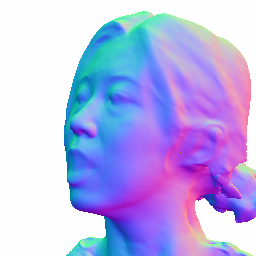}
&\includegraphics[height=\height]{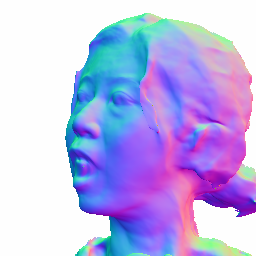}
& 
\\
D-Net & B-Morph & Fwd-Skin & C-Net & NerFACE~\cite{Gafni21} & Ours- & \textbf{Ours}& GT &
\end{tabularx}
\end{center}
\end{minipage}
\caption{
\textbf{Qualitative comparison on real data.} 
Unlike the baselines, \emph{Ours} generates complete images and accurate geometry even when extrapolating expressions beyond the training data. The examples become more challenging from top to bottom. 
\label{fig:real_qualitative}
}
\end{figure*}

To evaluate performance on real data, we compare to all baselines and NerFACE~\cite{Gafni21}. Additional comparisons with Varitex~\cite{buehler2021varitex}, Zhakarov et al.~\cite{zakharov2020fast} and HyperNeRF\cite{Park21b} can be found in \suppmat. 
Figure \ref{fig:real_qualitative} shows that all methods can generate realistic and correctly-posed images for easy expressions (row 1 and 2).
For easy expressions, Ours, Ours- and Backward Morphing (B-Morph) achieve the most accurate geometry because they leverage the morphing formulation of the FLAME face model. The FLAME-guided morphing field enables the joint optimization of a single canonical representation using information from all frames, even though they contain different expressions and poses. In contrast, C-Net, Fwd-Skin, and NerFACE~\cite{Gafni21} model expression deformations via direct conditioning on the implicit fields, and do not enforce a shared canonical geometry for different expressions. These methods are too under-constrained to deal with the ill-posed problem of extracting geometry from monocular videos. D-Net, on the other hand, does model the correspondence between frames. However, given the complexity of expression and pose deformations, the displacement-based deformation network cannot accurately represent the warping field, which leads to reduced quality.
Inspecting Fig.~\ref{fig:real_qualitative} top-to-bottom, reveals that the performance of the baselines degrades with increasingly strong poses and expressions.
For the last two rows, we show strong extrapolations where only our method can generate reasonable outputs that faithfully reflect the expression and pose conditions.
This finding is also supported by Tab.~\ref{tab:sota}. 
When evaluating on test videos with novel expressions and poses, our method achieves a much lower facial keypoint error, indicating more accurate reconstruction of expressions. Ours- achieves similar image and geometry quality as our full model, but is not as accurate in expression deformations. 
We show in \suppmat.~that the FLAME loss can be replaced via more training data without losing accuracy. 

Figure \ref{fig:real_interpolate_extrapolate}, demonstrates control over expressions and poses by interpolating and extrapolating an example expression (first expression component in FLAME~\cite{Li17}), plus jaw (pitch), and neck (yaw) poses separately. For each parameter, we show generated images and the training data distribution with 5 vertical lines corresponding to the 5 samples. 
This shows that our method generalizes to expressions and poses far beyond the training distribution. More interpolation and extrapolation examples can be found in \suppmat.
\begin{table}
\centering
\resizebox{\linewidth}{!}{ %
\small
\begin{tabular}{l|ccccc}
\toprule
Method & Expression $\downarrow$& $L_1$ $\downarrow$ & PSNR $\uparrow$ & SSIM $\uparrow$ & LPIPS $\downarrow$ \\
\midrule
C-Net & 3.615 & 0.05824 & 22.23 & 0.9524 & 0.03421\\
D-Net & 3.769 & 0.06130 & 21.77 & 0.9474 & 0.03227\\
B-Morph & 2.786 & 0.04980 & 23.50 & 0.9599 & 0.02231\\
Fwd-Skin & 3.088 & 0.05456 & 22.92 & 0.9586 & 0.02781\\
NerFACE~\cite{Gafni21} & 2.994 & \textbf{0.04564} & 23.58 & 0.9596 & 0.02156\\
Ours- & 2.843 & 0.04918 & 23.68 & 0.9615 & 0.02155\\
\textbf{Ours} & \textbf{2.548} & 0.04878 & \textbf{23.91} & \textbf{0.9655} & \textbf{0.02085}\\

\bottomrule
\end{tabular}
} 
\caption{
\textbf{Quantitative results on real videos.} We compare our method with the SOTA and baselines on test sequences with unseen expressions and poses.
Our method reconstructs the expressions more accurately while being on par in terms of image quality.
} %

\label{tab:sota}
\end{table}

\section{Conclusion}

We propose \methodname{}, an implicit morphable head avatar, controlled via expression and pose parameters in a similar way to 3DMMs, but with the ability to model diverse and detailed hairstyles and facial appearance. Our method---learned end-to-end from RGB videos---demonstrates accurate deforming geometry and extrapolates to strong expressions beyond the training distribution. 

While our method contributes towards building controllable implicit facial avatars, some challenges remain. 
First, surface representations achieve detailed facial geometry, but they cannot model the fine occlusions produced by hair. Future work could address this by combining volumetric representations~\cite{Mildenhall20} with animatable surfaces.
Second, the iterative non-rigid ray marching makes IMavatar slow to train ($\sim2$ GPU days). Initializing with mesh-ray intersections could speed up the process, as done in concurrent work~\cite{Jiang22}.
Third, our method relies on accurate face tracking and our performance degenerates with noisy 3DMM parameters (See \suppmat.). Refining the poses and expressions during training is a promising future direction. 
Finally, the appearance in the mouth interior region can be unrealistic (last two examples in Fig.~\ref{fig:real_qualitative}). We propose an improvement in the \suppmat.
We discuss potential negative societal impact in  light of disinformation and 
deep-fakes in the \suppmat.

{
\boldparagraph{Acknowledgements}
We thank the authors of \cite{Grassal21} for sharing their evaluation dataset. Yufeng Zheng and Xu Chen were supported by the Max Planck ETH Center for Learning Systems. This project has received funding from the European Research Council (ERC) under the European Union’s Horizon 2020 research and innovation program grant agreement No 717054.
MJB has received research gift funds from Adobe, Intel, Nvidia, Meta/Facebook, and Amazon.  MJB has financial interests in Amazon, Datagen Technologies, and Meshcapade GmbH.  MJB's research was performed solely at, and funded solely by, the Max Planck.
}

\onecolumn
\begin{center}
\textbf{\large Supplemental Materials}
\end{center}
\setcounter{equation}{0}
\setcounter{figure}{0}
\setcounter{table}{0}
\setcounter{page}{1}
\setcounter{section}{0}

This supplementary document provides additional ablation studies and results in Sec.~\ref{sec:supp_results}, implementation and training details in Sec.~\ref{sec:supp_details}, and discussion on broader impact in Sec.~\ref{sec:broader_impact}. Please also watch the accompanying video to see animated results and hear an explanation of our proposed method.

\section{Additional Ablations and Results}
\label{sec:supp_results}
\subsection{Comparison with Additional SOTAs}
Tab.~\ref{tab:additional_sota} lists comparisons with additional SOTA methods where code is publicly available. We run the pretrained models for Zhakarov et al.~\cite{zakharov2020fast} and Buehler et al.~\cite{buehler2021varitex} and train HyperNeRF~\cite{Park21b} for all four real subjects. Since HyperNeRF is not 3DMM-controllable, we condition the warping and slicing networks on the FLAME expression and pose parameters instead of a learnable latent code $\omega_i$.

\vskip -0.3em
\begin{table}[h!]
\centering
\resizebox{0.6\linewidth}{!}{
 \begin{tabular}{lccccc}
\toprule
Method & Expression $\downarrow$& $L_1$ $\downarrow$ & PSNR $\uparrow$ & SSIM $\uparrow$ & LPIPS $\downarrow$ \\
\midrule
Zhakarov et al. & 17.107 &	0.13929 &	15.24 &	0.8900 &	0.07040\\
VariTex & 3.704	& 0.09968 &	17.01 &	0.9233 &	0.04890\\
HyperNeRF & 7.201 &	0.08143 &	18.94 &	0.9207 &	0.03953 \\
\textbf{Ours} & \textbf{2.548} & \textbf{0.04878} & \textbf{23.91} & \textbf{0.9655} & \textbf{0.02085}\\
\bottomrule
\end{tabular}
}
\vspace{-0.6em}
\caption{\textbf{Additional SOTA baselines.} We evaluate SOTA baselines on our real dataset, and provide quantitative comparisons.
\label{tab:additional_sota}
}
\end{table}
\vskip -0.5em
\subsection{Experiment on MakeHuman Synthetic Dataset~\cite{Grassal21}}
\begin{table}[h!]
\vskip -0.2cm
\centering

\resizebox{0.6\linewidth}{!}{
     \begin{tabular}{lcccc}
     \toprule
     Metric & Female 1 & Female 2 & Male 1 & Male 2\\
     \midrule
     $\uparrow$ Mask IoU & 0.972& 0.973& 0.966&0.971 \\
     $\downarrow$ RGB L1(Intersec) &0.035 &0.025 &0.019 &0.039 \\
     $\uparrow$ Normal(Intersec) & \textbf{0.961}& \textbf{0.966}& \textbf{0.954}& \textbf{0.955}\\
     \midrule
     \cite{Grassal21} on Normal(Intersec) & 0.94& 0.95& 0.94& 0.94\\
     \bottomrule
     \end{tabular}
}
\caption{\textbf{MakeHuman.} IMavatar is competitive with concurrent work~\cite{Grassal21} without test-time pose optimization.
\label{tab:makehuman}
}
\end{table}
\begin{figure}[!htb]
\begin{center}
\setlength\tabcolsep{1pt}
\includegraphics[width=0.8\linewidth]{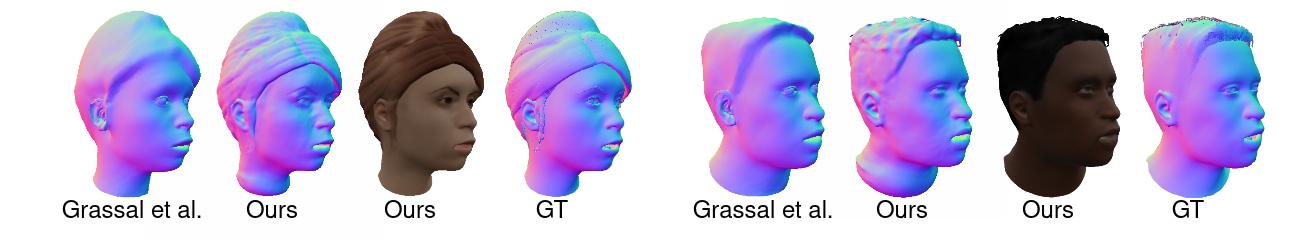}
\caption{\textbf{MakeHuman.} IMavatar learns accurate and detailed deformable geometry from monocular RGB videos.%
\label{fig:makehuman}
}
\end{center}
\end{figure}
We follow Grassal et al.~\cite{Grassal21} and train with the same frames for 4 subjects from MakeHuman. Tab.~\ref{tab:makehuman} lists IoU between the predicted and GT masks, image L1 over the intersection, and compares geometry with~\cite{Grassal21}. Our method achieves more accurate geometry \emph{without} test-time pose optimization used in~\cite{Grassal21}. Qualitative results are shown in Fig.~\ref{fig:makehuman}. 

\subsection{Ablation on FLAME Pseudo GT Supervision}
\begin{figure}[ht!]
\centering
     \centering
     \includegraphics[width=0.8\linewidth]{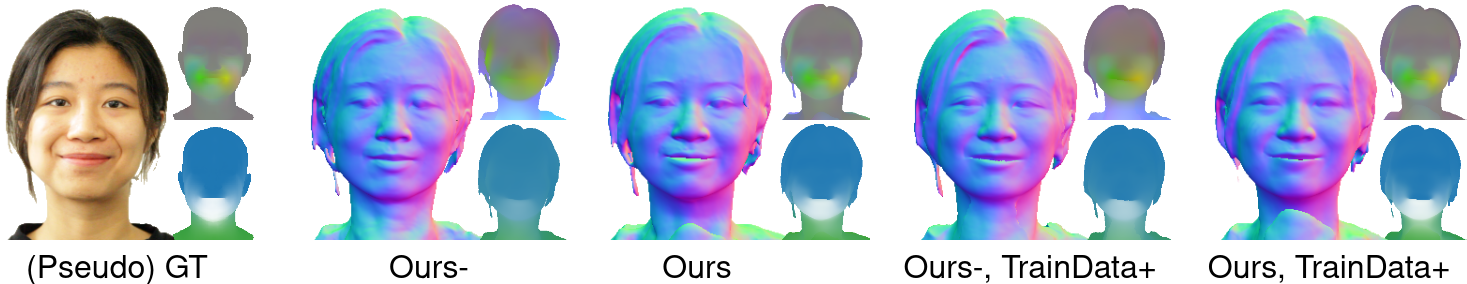}
\caption{Our method can be trained without FLAME pseudo ground-truth supervision. With more diversed training data, IMavatar learns more detailed expression and pose deformations. Neck geometry, however, is not guaranteed to be correct due to the lack of movement and ambiguity between head and neck rotation in the training data.
\label{fig:train_objectives}}
\end{figure}

Our method can also be trained \textit{without} 3DMM supervision, using only mask and RGB losses (`Ours-' in Tab.~\ref{tab:ablation_objectives} and Fig.~\ref{fig:train_objectives}). Expression error is higher without pseudo GT supervision (row 1 and 2). However, with TrainData+, which contains 30\% more frames and more expression variation than the original trainset, Ours- achieves comparable performance without leveraging pseudo GT (row 3 and 4). 
\begin{table}[ht!]
\centering
\resizebox{0.6\linewidth}{!}{
    \begin{tabular}{c|c|ccccc}
    \toprule
    Method & TrainData+ & Expr. $\downarrow$& $L_1$ $\downarrow$ & PSNR $\uparrow$ & SSIM $\uparrow$ & LPIPS $\downarrow$ \\
    \midrule
    Ours-& &  3.337& 0.0496& 22.32& 0.9532& 0.0317\\
    Ours& &  \textbf{2.973} & \textbf{0.0480}& \textbf{22.55}& \textbf{0.9572}& \textbf{0.0292}\\
    \midrule
    Ours-&x&  2.955& \textbf{0.0447}& 21.98& 0.9591& \textbf{0.0277}\\
    Ours&x&  \textbf{2.876}& 0.0457& \textbf{22.07}& \textbf{0.9596}& 0.0296\\
    \bottomrule
    \end{tabular}
}
\caption{\textbf{Pseudo GT supervision} improves metrics given limited data. With TrainData+, it is sufficient to learn unsupervisedly from images. Scores are calculated on one subject. See Fig.~\ref{fig:train_objectives} for qualitative results.
\label{tab:ablation_objectives}
}
\end{table}

\subsection{Mouth Interior Improvement}
\begin{figure}[ht]
     \centering
     \includegraphics[width=0.2\linewidth]{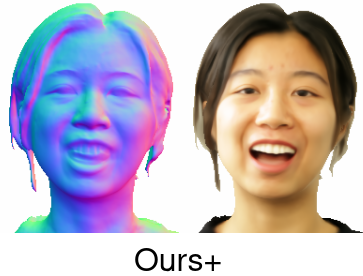}
     \caption{With more diverse data and pre-estimated semantic segmentations, IMavatar can learn better mouth interior geometry and texture.\label{fig:mouth_interior}}
\end{figure}

Simply training on a dataset with more expressions (TrainData+) and setting blendshape supervision in the mouth interior region to 0 (with estimated semantic maps) faithfully reproduces teeth (Fig.~\ref{fig:mouth_interior})

\subsection{Ablation on Pre-processing: Tracking and Segmentation}
We experiment on Female 2 from MakeHuman.

\paragraph{3DMM tracking:} We add uniformly distributed noise to the fitted 3DMM global, neck and jaw poses, with a noise range from 0.025(~1.4$^{\circ}$) to 0.1(~5.7$^{\circ}$).

\paragraph{Foreground mask:} We randomly select a $61\times61$ square and set the mask value to True or False randomly. We degrade 10\%, 50\% and 100\% of the masks. For both ablation experiments, the random degradation is a pre-processing step (not changed during training). See Tab.~\ref{tab:input_ablation}.
\begin{table}[h!]
\centering
\resizebox{0.6\linewidth}{!}{
    \begin{tabular}{l|ccc|c|ccc}
    \toprule
    Metric & \multicolumn{3}{c}{3DMM tracking}& Baseline& \multicolumn{3}{c}{Foreground Mask}\\
     &$0.1$&$0.05$ & $0.025$ & & $10\%$ &$50\%$ &$100\%$ \\
    \midrule
    $\uparrow$ Mask IoU & 0.927& 0.954 &0.967 & \textbf{0.983}&0.982 & 0.980 & 0.975\\
    $\uparrow$ Normal& 0.905 &0.938&0.959& 0.958 & \textbf{0.961} & 0.958 &0.932\\
    $\downarrow$ RGB L1 &0.062& 0.045& 0.032&\textbf{0.023} &0.024 & 0.027 &0.029\\
    \bottomrule
    \end{tabular}
}
\caption{\textbf{Ablation Pre-processing.} IMavatar relies on accurate 3DMM tracking, but it is reasonably  robust to mis-segmentations. \label{tab:input_ablation}}
\end{table}
\begin{figure}[ht]
\begin{center}
\setlength\tabcolsep{1pt}
\includegraphics[width=0.6\linewidth]{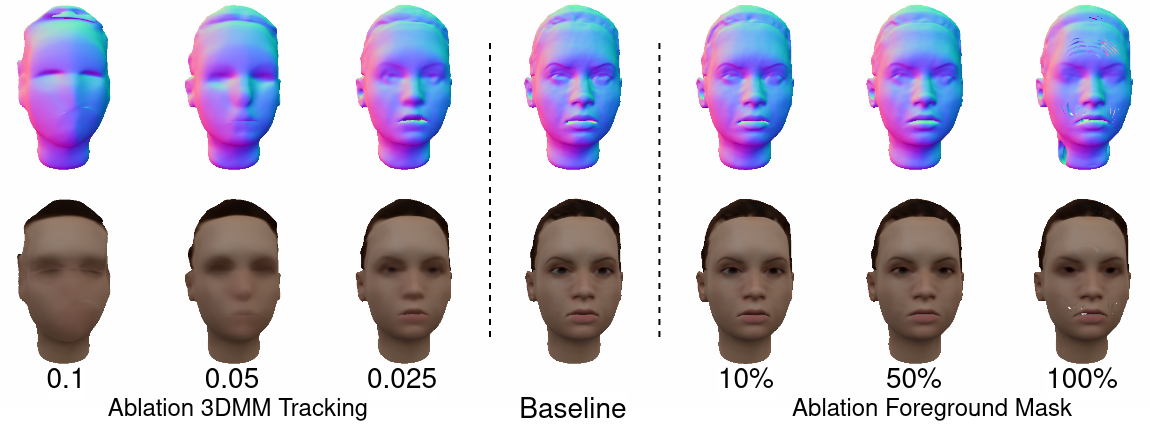}
\caption{\textbf{Ablation Pre-processing.} Noisy 3DMM tracking is a major reason for blurry texture and geometry. Applying degradation to all masks leads to visible artifacts.
\label{fig:input_ablation}
}
\end{center}
\end{figure}
\subsection{Jaw Pose Extrapolation}
\begin{figure}[]
\centerline{
\includegraphics[width=0.75\linewidth]{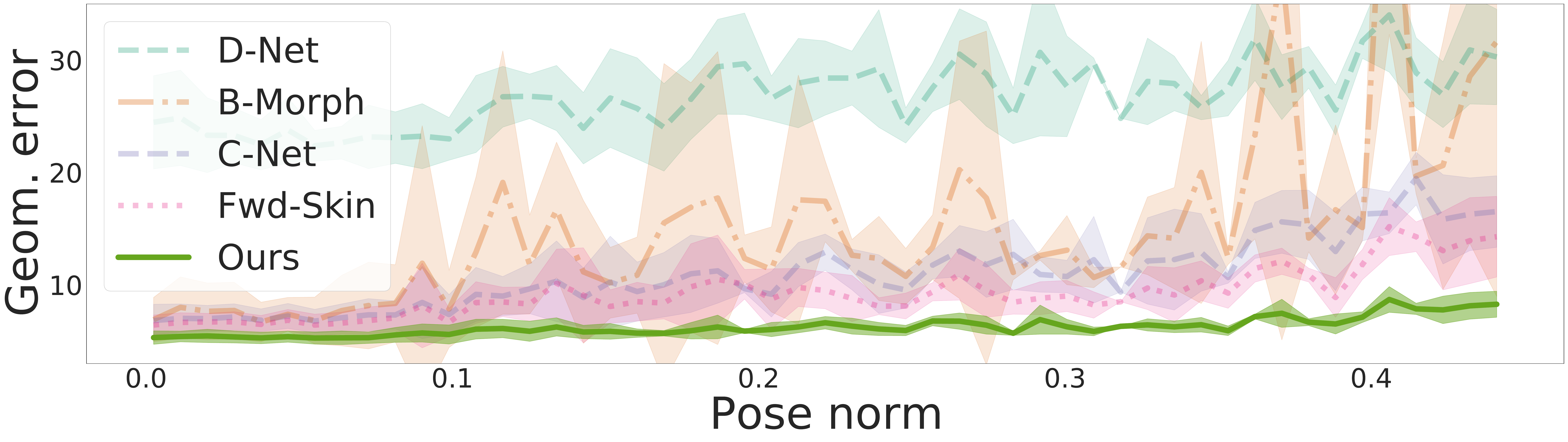}
}
\caption{
\textbf{Jaw pose extrapolation.} The x-axis denotes the norm of the jaw pose parameters in radians. The y-axis plots 
the angular error of the surface normals (lower is better).%
 Performance of baseline methods worsen drastically as pose become more extreme. 
}
\label{fig:supp_pose_vs_norm}
\end{figure}

Fig. 4 in the main paper shows how the performance of baseline methods drops for stronger expressions. We extend this comparison by plotting the error with respect to the norm of the jaw pose parameter in Fig.~\ref{fig:supp_pose_vs_norm}. 
Our method achieves low errors even for strong jaw poses, while the error for baseline methods increases drastically. %

\subsection{Additional Extrapolation Results}
We show extrapolations as animations in our supplementary video. For each expression, we interpolate the individual FLAME expression parameter from [-4, 4], and keep all other pose and expression parameters fixed as zero. We show the smiling (1st), lip side movement (3rd), and eyebrow raising (10th) expressions.

\section{Implementation Details}
\label{sec:supp_details}
\subsection{Mesh Morphing v.s. Implicit Morphing}
\begin{figure}[t]
\begin{center}
\includegraphics[width=0.9\textwidth]{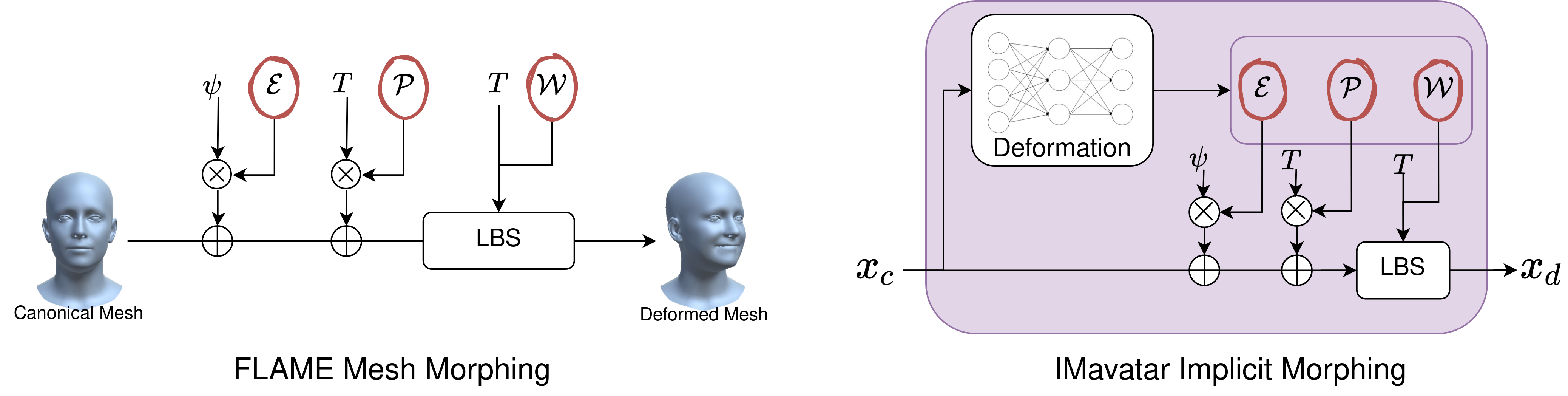}\\

\end{center}
\caption{
\textbf{FLAME morphing v.s. Implicit Morphing.} }
\label{fig:FLAME_implicit morphing}
\end{figure}

A visual illustration of mesh morphing versus implicit morphing can be found in Fig.~\ref{fig:FLAME_implicit morphing}
\subsection{Network Architecture}
\begin{figure}[t]
\begin{center}
\includegraphics[width=0.9\textwidth]{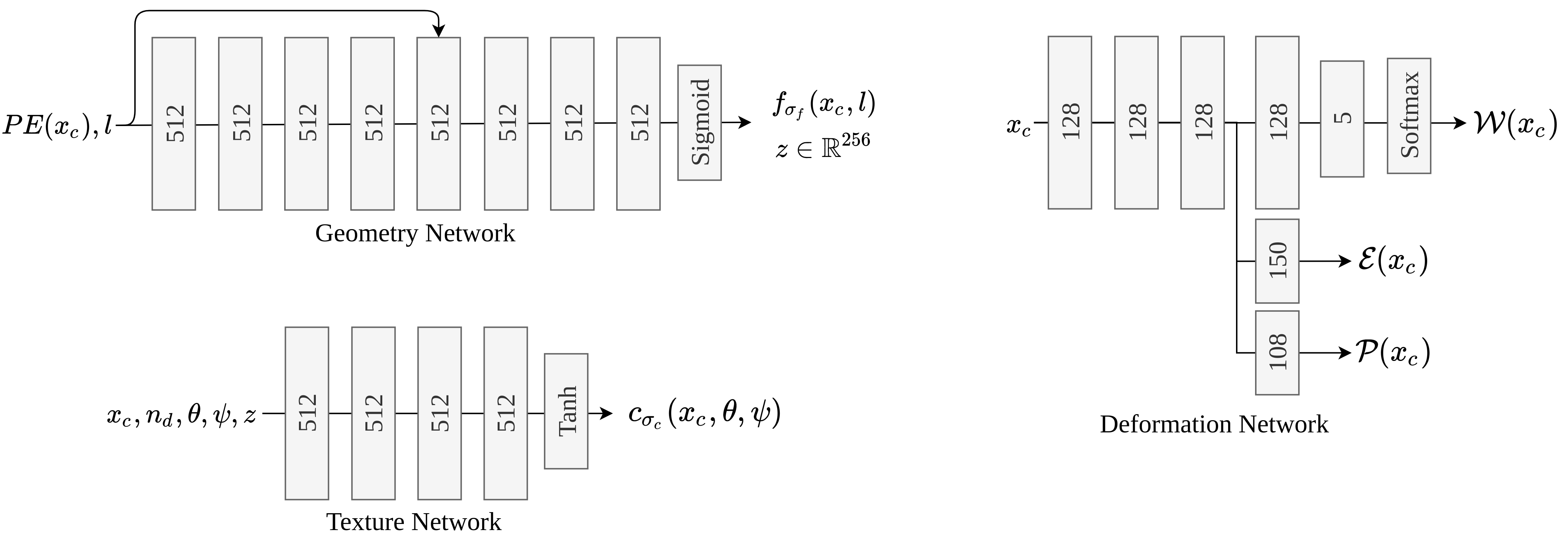}\\

\end{center}
\caption{
\textbf{Network Architecture.} Each block represents a linear layer with its output dimension specified in the inset, followed by a weight normalization layer~\cite{Salimans16} and an activation layer. We use Softplus~\cite{Dugas00} activation for the geometry and deformation network, and ReLU activation for the texture network. $z \in \mathbb{R}^{256}$ is the latent feature from the geometry network which is used as an input condition for the texture network. 
}
\label{fig:supp_network_architecture}
\end{figure}
\begin{figure}[t]
\begin{center}
\includegraphics[width=0.9\textwidth]{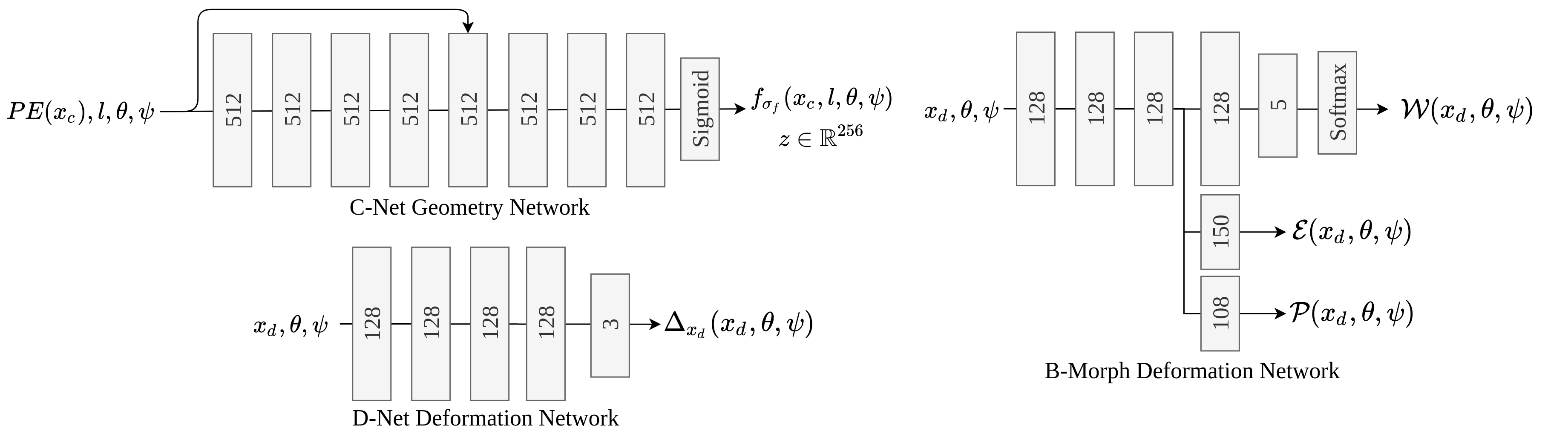}\\

\end{center}
\caption{
\textbf{Network Architecture for Baselines.} We show the modified geometry network for C-Net, which is additionally conditioned on the expression and pose parameters, $\psi$ and $\theta$. The deformation network for the B-Morph baseline is conditioned on the deformed point $x_d$ and the expression and pose parameters. For D-Net, the input condition is the same as B-Morph, but the output is the displacement distance for the deformed location.
}
\label{fig:supp_ablation_architecture}
\end{figure}
We implement our models in PyTorch~\cite{Paszke19}. The network architectures for the geometry-, texture-, and deformation-networks are illustrated in Fig.~\ref{fig:supp_network_architecture}. We initialize the geometry network with geometric initialization~\cite{Atzmon20} to produce a sphere at the origin. For the deformation network, we initialize the linear blend skinning weights to have uniform weights for all bones and the expression and pose blendshapes to be zero. For the geometry network, we use positional encoding~\cite{Mildenhall20} with 6 frequency components, and condition on a per-frame learnable latent code $\boldsymbol{l} \in \mathbb{R}^{32}$. 

Fig.~\ref{fig:supp_ablation_architecture} shows the modified geometry network for the C-Net, and the deformation network for the D-Net and B-Morph baselines (see Section 4.2 in the main paper for definitions). We initialize the ablated geometry and deformation networks in the same way as our method. The displacement output for D-Net is also initialized to be zero.

\subsection{Ray Tracing}
Our ray tracing algorithm is similar to IDR~\cite{Yariv20}, except that we do not perform the sphere ray tracing with signed distance values (SDF). This is because SDFs are not guaranteed to be correct in value after non-rigid deformation, and might lead to over-shooting. For this reason, we also eliminated the Eikonal loss~\cite{Gropp20} in IDR~\cite{Yariv20}, and reconstruct an occupancy field instead of an SDF field. 

\subsection{Correspondence Search}
Following SNARF~\cite{Chen21}, for each deformed point $x_d$ we initialize the canonical point $x_c$ in multiple locations. More specifically, we inversely transform $x_d$ with the transformation matrix of the head, jaw, and shoulder to ensure one of the initialized locations is close enough to the canonical correspondence. Then, we leverage Broyden's method~\cite{Broyden65} to find the root of $w_{\sigma_d}(x_c) = x_d$ in an iterative manner. We set the maximum number of update steps to $10$ and the convergence threshold to $1e-5$. In the case of multiple converged canonical correspondences, the occupancy of the deformed point is defined as the minimum of all occupancy values.

\subsection{Canonical Pose}
The canonical pose of FLAME~\cite{Li17} has neutral expressions $\psi=0$ and poses $\theta=0$ (mouth closed). This results in a sharp transition in skinning weights and blendshape values between the lips. We modified the pitch value of the jaw to $0.2$ to slightly open the mouth, in order to avoid the sharp boundary between the lips which is difficult to approximate with coordinate-based MLPs. 

\subsection{Training Details}
We train our network for 60 epochs with Adam optimizer~\cite{kingma14} using a learning rate of $\eta=1e^{-4}$, and $\beta = (0.9, 0.999)$. Learning rate is decayed by $0.5$ after 40 epochs. 

\subsection{Real video dataset pre-processing}
Our training and testing videos are all captured with one single fixed camera. For training, we record two videos: one head rotation video to capture the full facial appearance from different angles, and one talking video to capture common and mild expressions in a speech sequence. For testing, we ask the subjects to perform strong unseen expressions such as a big smile, jaw opening, pouting, and rising of the eyebrows.

For both training and testing videos, we use DECA~\cite{Feng21} to regress the initial FLAME~\cite{Li17} shape, expression, and pose parameters. Unfortunately, the eye poses (gaze directions) are not tracked in our pre-processing pipeline. To refine the regressed FLAME parameters, we estimate the facial keypoint with~\cite{Bulat17} and optimize the regressed parameters and global translation vectors jointly. The primary optimization objective is the keypoint error:
\begin{equation*}
    E_{kp} = \left\|K(\theta, \psi, \beta, t) - K^{target}\right\|_2,
\end{equation*}
where $K(\theta, \psi, \beta, t)$ are the predicted 2D keypoints from FLAME pose, expression and shape parameters $\theta, \psi, \beta$ and global translation $t$, and $K^{target}$ are the optimization targets predicted by~\cite{Bulat17}. We use one single shape parameter $\beta$ for each subject. During optimization, we regularize shape and expression by:
\begin{equation*}
    E_{reg} = \lambda_{\beta}\left\|\beta\right\|_2^2 +  \frac{\lambda_{\psi}}{T} \sum_{\tau \in [0, T]}\left\|\psi_{\tau}\right\|_2^2,
\end{equation*}
where $\beta$ and $\psi$ are the shape and expression parameters, $T$ is the number of frames, and $\lambda_{\beta}$ and $\lambda_{\psi}$ are objective weights, set to $1e-4$ and $2e-4$, respectively. We also leverage temporal consistency terms for expression, pose and global translations:
\begin{equation}
    E_{temp} =\frac{1}{T-1} \sum_{\tau \in [0, T-1]}
    (\lambda_{\psi}^{temp}\left\|\psi_{\tau+1}  - \psi_{\tau}\right\|_2^2 
    + \lambda_{\theta}^{temp}\left\|\theta_{\tau+1}  - \theta_{\tau}\right\|_2^2 
    + \lambda_{t}^{temp}\left\|t_{\tau+1}  - t_{\tau}\right\|_2^2),
\end{equation}
where $\theta$ and $t$ are the pose parameters and global translation vectors. $\lambda_{\psi}^{temp}$, $\lambda_{\theta}^{temp}$ and $\lambda_{t}^{temp}$ are set to $1e-3$, $\frac{2}{3}$ and $\frac{10}{3}$, respectively. The final optimization objective can be represented as:
\begin{equation*}
    E = E_{kp} + E_{reg} + E_{temp}.
\end{equation*}
We will release the pre-processing pipeline for real videos.
\section{Broader Impact}
\label{sec:broader_impact}
Our work reconstructs a high fidelity facial avatar from monocular videos, which can extrapolate to unseen expressions given a training video of only mild deformations. This takes an important step towards democratizing 3D acquisition devices, as it does not require the user to have access to expensive capture equipment in order to get an animatable 3D model of itself. Thanks to the extrapolation abilities, it does not impose overly restrictive constraints in terms of the capture process itself, greatly simplifying it without sacrificing in geometric quality. 

There is nevertheless the danger of nefarious use of any technology that can generate plausible renderings of individuals under fine grained control of expressions and head pose. The foremost danger here is the use of so-called deep-fakes and dispossession of identity. We are aware of the potential for abuse of our technology -- despite its intended use for positive causes such as connecting people via mixed reality videoconferencing. We argue that performing research on topics such as this are best performed in an open and transparent way, including full disclosure of the algorithmic details, data and models which we intend to release for research purposes. While we may, unfortunately, not be able to prevent the development of deep-fake technologies entirely, we may however inform the general understanding of the underlying technologies and we hope that our paper will therefore also be useful to inform counter-measures to nefarious uses. %

{
    \small
    \bibliographystyle{ieee_fullname}
    \bibliography{macros,main}
}



\end{document}